%% file: main.tex
\documentclass[10pt,twocolumn,letterpaper]{article}

\usepackage[pagenumbers]{cvpr} 

\usepackage{amsmath}
\usepackage{amssymb}
\usepackage{bm}
\usepackage{bbm}
\usepackage{acro}
\usepackage{multicol}
\usepackage{multirow}
\usepackage{siunitx}

\input{sec/notations}

\input{sec/macros}
\input{sec/acronyms}

\definecolor{cvprblue}{rgb}{0.21,0.49,0.74}
\usepackage[pagebackref,breaklinks,colorlinks,allcolors=cvprblue]{hyperref}

\def\confName{CVPR}
\def\confYear{2027}

\title{Auditing Adversarial Robustness with Training-time Canaries}

\author{Yoojin Jung\\
Aalto University\\
Espoo, Finland\\
{\tt\small yoojin.jung@aalto.fi}
\and
Sebastian Szyller\\
Aalto University\\
Espoo, Finland\\
{\tt\small contact@sebszyller.com}
}

\begin{document}
\maketitle
\input{sec/0_abstract}    
\input{sec/1_intro}
\input{sec/2_background}

\input{sec/3_problem}

\input{sec/4_method}
\input{sec/5_eval}

\input{sec/6_discussion}
\input{sec/7_related}
\input{sec/8_conclusion}
\input{sec/9_LLM}
{
    \small
    \bibliographystyle{ieeenat_fullname}
    \bibliography{references}
}

\input{sec/X_suppl}

\end{document}

%% file: sec/notations.tex
\def\gL{{\mathcal{L}}}

\def\vtheta{{\bm{\theta}}}

\def\vx{{\bm{x}}}

\DeclareMathAlphabet{\mathsfit}{\encodingdefault}{\sfdefault}{m}{sl}
\SetMathAlphabet{\mathsfit}{bold}{\encodingdefault}{\sfdefault}{bx}{n}

\def\gC{{\mathcal{C}}}

\def\gL{{\mathcal{L}}}

\def\gX{{\mathcal{X}}}
\def\gY{{\mathcal{Y}}}

\newcommand{\pdata}{p_{\rm{data}}}



%% file: sec/macros.tex
\newcommand{\adv}{\ensuremath{\mathcal{A}}\xspace}

%% file: sec/acronyms.tex
\DeclareAcronym{ml}{short  = ML,  long = machine learning}
\DeclareAcronym{asr}{short = ASR, long = attack success rate}
\DeclareAcronym{pgd}{short = PGD, long = projected gradient descent}
\DeclareAcronym{aa}{short = AA, long = auto-attack}
\DeclareAcronym{fgsm}{short = FGSM, long = fast gradient signed method}
\DeclareAcronym{dp}{short = DP, long = differentiable privacy}
\DeclareAcronym{ifr}{short = IFR, long = initially flipped rate}
\DeclareAcronym{ctr}{short = CTR, long = conditional targeted recovery}
\DeclareAcronym{at}{short = AT, long = adversarial training}
\DeclareAcronym{rs}{short = RS, long = radonmized smoothing}
\DeclareAcronym{trades}{short = TRADES, long = TRADES}
\DeclareAcronym{mart}{short = MART, long = MART}
\DeclareAcronym{crown}{short = CROWN, long = CROWN}
\DeclareAcronym{ibp}{short = IBP, long = interval bound propagation}
\DeclareAcronym{mrr}{short = MRR, long = Mean Recovery Radius}

%% file: sec/0_abstract.tex
\begin{abstract}
Adversarial robustness is commonly evaluated using empirical robust accuracy against specific attacks or certified robustness guarantees.
However, such evaluations focus on measuring defense performance on fixed data, 
rather than deliberately constructing cases that may expose weaknesses in the defense. 
To complement existing robustness evaluation, we propose a canary-based adversarial robustness auditing framework designed to elicit worst-case examples.
Our method constructs training-time canaries and uses feedback from the trained model to move them toward its decision boundary;
it produces stronger auditing probes that higher attack success rate than ``normal'' adversarial examples and typically at substantially lower $\epsilon$ values.
We evaluate our framework on multiple empirical and certified defenses by comparing standard robustness evaluation with canary-based auditing.
While standard robust accuracy remains largely unchanged, our audit reveals that across the tested settings, the canaries are easier to attack,
and typically with smaller perturbations.
These findings demonstrate that systematic failures may remain unobserved under standard robustness evaluation 
and motivate worst-case auditing as an important part of robustness assessment.
\end{abstract}

%% file: sec/1_intro.tex
\section{Introduction}
\label{sec:intro}

\Ac{ml} models are susceptible to various attacks.
An \emph{adversarial example} is a deliberately perturbed input, crafted to cause a misclassification while remaining innocuous to a human inspector~\cite{goodfellow2015explainingharnessingadversarial,szegedy_intriguing_2013,elsayed_adversarial_2018}.
\emph{Adversarial robustness} is the model's ability to keep consistent predictions on such inputs.
Defenses can be divided into two categories: \emph{empirical robustness} (e.g. adversarial training~\cite{madry2018deeplearningmodels}) and \emph{certified robustness} (e.g. randomized smoothing~\cite{cohen2019certifiedadversarialrobustness}).

Empirical robustness is typically evaluated using \ac{asr} -- fraction of initially correctly classified inputs for which the attack succeeds, and robust accuracy -- fraction of the full test set that remains correctly classified under attacks~\cite{croce2020reliableevaluationadversarial,dong_benchmarking_2020,su_is_2018}.
On the other hand, certified robustness provides \emph{certified radius}
within which there does not exist a viable adversarial example~\cite{cohen2019certifiedadversarialrobustness,wong_provable_2018,gowal2019effectivenessintervalbound}.

A substantial part of adversarial defense evaluation has focused on ``breaking'' proposed defenses, through stronger or adaptive attacks~\cite{tramer2020adaptiveattacksadversarial,wang_diffhammer_2024,zhou_vulnerability_2024}, simple reparameterization of the attack~\cite{kannan2018adversariallogitpairing,engstrom2018evaluatingunderstandingrobustness}, or by highlighting methodological flaws~\cite{athalye2018obfuscatedgradientsgive,carlini2024cuttingbuggyadversarial}. 
Other work has broadened evaluation through standardized benchmarks~\cite{croce2021robustbenchstandardizedadversarial}, multiple attack types and strengths~\cite{dai_multirobustbench_2023}, and distribution shifts~\cite{li_oodrobustbench_2023}.
More recent work has questioned whether aggregate or fixed-budget robustness metrics adequately characterize model behavior~\cite{dong_benchmarking_2020,olivier_how_2023,cina_attackbench_2024}. This has motivated complementary analyses of per-example vulnerability and the distance to adversarial failure~\cite{pintor_fast_2021,han_characterizing_2023,kim_adversarys_2026}.

In this work, we argue that existing evaluations are too coarse and do not provide a complete overview of the robustness.
In turn, the generalization of the robustness claims is overstated, and hides the risk of the most susceptible samples, i.e., worst-case settings~\cite{han_characterizing_2023,ngnawe_detecting_2024,sun_average_2025}.
Inspired by a long line of work on auditing differentially private stochastic gradient descent~\cite{abadi2016deeplearningdifferential,jagielski2020auditingdifferentiallyprivate,steinke2023privacyauditingone,cebere2024tighterprivacyauditing,yaghini2026optifluenceprincipleddesign},
we introduce a \emph{canary-based}, training-time auditing framework for adversarial robustness.
We demonstrate that popular empirical defenses fail like this;
whereas certified defenses fail like.

In summary, we make the following contributions:
\begin{itemize}
\item We introduce a canary-based framework for adversarial robustness auditing that complements standard robustness evaluation with worst-case probes.

\item  We introduce a framework for auditing robustness with training-time canaries. The framework relies on logit-margin based canary selection, their embedding in the model, and targeted recovery procedure.

\item  We demonstrate the effectiveness of our framework across empirical and certified defenses, datasets, and architectures; the canary audit reveals systematic weaknesses not captured by standard robustness evaluation.

\end{itemize}

%% file: sec/2_background.tex
\section{Background}
\label{sec:background}

\begin{figure*}[t]
    \centering
    \includegraphics[width=\linewidth]{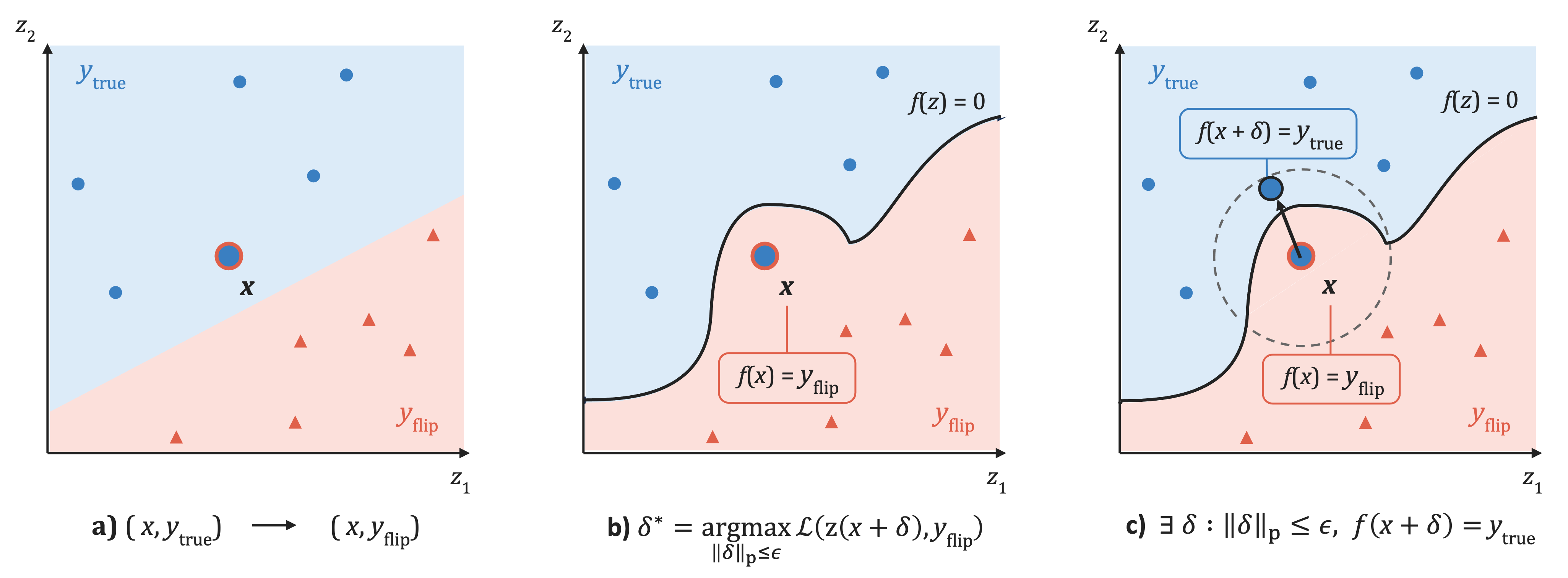}
    \caption{
    \textbf{Canary-based robustness auditing.}
    (a) A training example $(\vx,y_{\mathrm{true}})$ is selected according to a canary construction criterion, and its label is replaced with $y_{\mathrm{flip}}$ while $\vx$ remains unchanged.
    (b) Robust training proceeds with $(\vx,y_{\mathrm{flip}})$, resulting in a model that predicts the flipped label for the canary.
    (c) The audit searches for a targeted perturbation $\|\delta\|_p \le \epsilon$ that restores the original label $y_{\mathrm{true}}$.
    }
    \label{fig:1}
\end{figure*}

\subsection{Adversarial examples}\label{sec:att_def}\label{sec:a_example}
Let $f(\vx;\vtheta):\gX\to\gY$ denote a classifier with parameters $\vtheta$, where $\vx\in\gX$ is an input and $y\in\gY$ is its ground-truth label.
Let $\pdata$ denote the data-generating distribution over $\gX\times\gY$.
Adversarial examples are inputs deliberately perturbed to induce an incorrect prediction while remaining close to the original input under a specified perturbation measure~\cite{szegedy_intriguing_2013,goodfellow2015explainingharnessingadversarial,carlini_towards_2016}.
We consider additive perturbations:
\begin{equation}
    \delta : \|\delta\|_{p} \le \epsilon,
\end{equation}
where $p$ specifies the norm and $\epsilon$ bounds the perturbation magnitude.
Let $\vx'=\vx+\delta$ denote the perturbed input,
an untargeted adversarial example satisfies
\begin{equation}
    f(\vx';\vtheta) \neq y.
\end{equation}

Let $z(\vx;\vtheta)\in\mathbb{R}^{|\gY|}$ denote the logit vector associated with the classifier output.
A common white-box formulation searches for a perturbation that maximizes the loss with respect to the ground-truth label:
\begin{equation}
    \delta^{*}\in \arg\max_{\|\delta\|_p \le \epsilon}
    \gL\bigl(z(\vx + \delta;\vtheta), y\bigr).
\end{equation}

For a target label $y_{\mathrm{tgt}}\neq y$, a targeted attack instead seeks a perturbation such that 
\begin{equation}
    f(\vx';\vtheta)=y_{\mathrm{tgt}}.
\end{equation}

\subsection{Adversarial defenses}\label{sec:a_def}
Adversarial defenses aim to preserve correct model behavior under perturbations within a specified threat model.
In this work, we consider two classes of defenses: empirical and certified.

\noindent\textbf{Empirical robustness.}\label{sec:emp_rob}
Empirical defenses aim to improve model performance against adversarially perturbed inputs.
A canonical approach is \ac{at}~\cite{madry2018deeplearningmodels}, which minimizes the loss on training samples perturbed to maximize that loss within a prescribed perturbation budget:
\begin{equation}
\min_{\vtheta}
\mathbb{E}{(\vx,y)\sim\pdata}
\left[
\max{\|\delta\|_p\le\epsilon}
\gL\bigl(z(\vx+\delta;\vtheta),y\bigr)
\right].
\end{equation}
Subsequent methods modify this objective to improve the robustness--accuracy trade-off or emphasize difficult and misclassified examples, as in \Acl{trades}~\cite{zhang2019theoreticallyprincipledtradeoff} and \Acl{mart}~\cite{wang2020improvingadversarialrobustness}.
Since these methods do not provide formal guarantees, their robustness is assessed through attack-based evaluation, where stronger attacks provide stronger empirical evidence~\cite{carlini2019evaluatingadversarialrobustness,croce2020reliableevaluationadversarial}.

\noindent\textbf{Certified robustness.}\label{sec:cert_rob}
Certified defenses aim to provide a formal guarantee that the model's prediction remains unchanged for all perturbations within a specified region around an input:
\begin{equation}
f(\vx+\delta;\vtheta)=y
\qquad
\forall,\delta:\|\delta\|_p\le\epsilon.
\end{equation}
Common approaches include \ac{rs}~\cite{cohen2019certifiedadversarialrobustness}, which gives probabilistic certificates for a smoothed classifier, and \ac{ibp}~\cite{gowal2019effectivenessintervalbound}, which trains models with verifiable output bounds.
Unlike empirical robustness, which is established through observed attack outcomes, certified robustness associates individual inputs with robustness guarantees under the assumptions of the certification method.

%% file: sec/3_problem.tex
\section{Problem statement} \label{sec:problem}

\subsection{Threat model}\label{sec:threat}
We use the established white-box threat model for adversarial robustness.
The goal of the adversary \adv is to craft an adversarial example $\vx'$ that leads to an untargeted ($f(\vx';\vtheta) \neq y$) or targeted ($f(\vx';\vtheta) = y_{\mathrm{tgt}}$) misclassification.

\adv operates at inference time and has white-box access to the defended classifier, including its architecture, parameters, defense mechanism, loss function, and gradients.

\adv may modify an input only through an additive perturbation $\delta$ satisfying $\|\delta\|_p \le \epsilon$ defined in Section~\ref{sec:background}.
Thus, any $\vx'$ satisfies
$\vx'=\vx+\delta$ $(\|\delta\|_p\le\epsilon)$.
The adversary cannot modify the training data, training procedure, or learned model parameters.
For randomized or certified defenses, we assume knowledge of the corresponding defense and certification procedures.

In this work, we focus on white-box adversarial examples.
However, the threat model and our auditing framework could equally be used with black-box adversaries.

\subsection{Defense reliability claims}\label{sec:rob_claim}

We use \emph{defense reliability claim} to describe how a defense behaves within some specified threat model and evaluation setting.
The claim may be explicit, as in a certified guarantee, or implicit, as when a reported empirical robustness metric is used as evidence that a defense is reliable.
We distinguish between empirical and certified claims below because the evidence required to challenge them is different.

\noindent\textbf{Empirical robustness claims.}\label{sec:emp_claim}
An empirical robustness claim is supported by the empirical behavior of a defense subject to an evaluation procedure.
For example, an empirical defense may report robust accuracy for attacks constrained to $\|\delta\|_p \le \epsilon$.
Such a result provides empirical evidence that the defense increases robustness in the evaluated conditions but does not establish that all in-scope inputs are robust
or that no consistently reachable failure mode exists.
Unlike certified claims, empirical robustness claims cannot be falsified by a single adversarial example,
since individual failures are already permitted by the reported metric.

\noindent\textbf{Certified robustness claims.}\label{sec:cert_claim}
A certified robustness claim asserts that the prediction for a specific input remains unchanged under all perturbations satisfying the specified constraint.
For example, a certificate may guarantee that an input-label pair $(\vx,y)$ is correctly classified for every perturbation satisfying $\|\delta\|_p\le\epsilon$.
Unlike an empirical robustness claim, this can be falsified by a single counterexample within the certified radius that changes the prediction.

\subsection{Auditing robustness}
\label{sec:rob_audit}

\noindent\textbf{Limitations of standard robustness evaluation.}
Standard robustness evaluation empirically measures average case robustness for a data distribution, i.e. how attack effectiveness degrades when a defense is deployed.
However, it does not convey how effective a defense is in the worst case.
A statement ``\textit{90\% robust accuracy with $\epsilon=4/255$ under $\|\delta\|_{\infty}\le\epsilon$}'' says nothing about the samples that are the easiest to attack.
The purpose of the audit is to elicit such cases.

This resembles how membership privacy evaluation has evolved.
Early works used average attack success rate that underestimated the risk~\cite{shokri2017membershipinferenceattacks,yeom2018privacyriskmachine};
later work focused on targeting outlier records with high confidence~\cite{carlini2022membershipinferenceattacks,zarifzadeh2024lowcosthighpowermembership}.

In our setting, the auditor has white-box access to the model, training data and the robust training procedure.
The auditor can modify or inject new training samples but cannot modify anything else.
The resulting model is used for the robustness audit -- within the threat model of the attack.
\noindent\textbf{Objective.}
\label{sec:audit_objective}
The goal of the audit is to find samples that expose the weakest robustness behavior given the threat model.
For \emph{empirical robustness}, the audit looks for conditions where the defense is less robust than the standard evaluation implies.
For certified robustness, a valid counterexample within a certified radius contradicts the corresponding guarantee.
The audit has three requirements.
\begin{enumerate}[label=\textbf{R\arabic*}]
    \item\label{req:severity} \textbf{(severity)} measures how strongly a failure exposes weak robustness behavior. For norm-bounded robustness, a failure that appears at a smaller perturbation magnitude is more severe.
    \item\label{req:coverage} \textbf{(coverage)}  captures how broadly such failures occur across audit probes. One audit may expose a few severe failures, while another exposes more failures at larger perturbation magnitudes. We therefore consider severity and coverage separately rather than reducing them to a single score.
    \item\label{req:cost} \textbf{(computational cost)} is a secondary consideration that we do not evaluate in this work. When two audits provide similar evidence, lower training and search costs are preferable.
\end{enumerate}

%% file: sec/4_method.tex
\section{Canary-based robustness auditing}
\label{sec:method}

The audit consists of constructing canaries before training, 
and, after training, testing whether the flipped behavior learned on these canaries can be reversed by perturbations allowed under the threat model.
We further consider an adaptive procedure that refines canaries using information obtained from a previously trained model.

\subsection{Training-time canaries}\label{sec:tt_canary}

Given a training example $(\vx, y)$,
we construct a canary by assigning the same input a different label
$\tilde{y} \in \gY \setminus \{y\}$:
\begin{equation}
    (\vx, y)
    \longrightarrow
    (\vx, \tilde{y}).
\end{equation}
The original example $(\vx, y)$ is replaced, not retained alongside the canary, $\vx$ remains unchanged.

After training, we check whether the resulting model has learned $\tilde{y}$ and predicts it:
\begin{equation}\label{eq:learned_canary}
    f(\vx;\vtheta) = \tilde{y}.
\end{equation}
Canaries that satisfy this condition are used in the recovery audit described in Section~\ref{sec:recovery_audit}.
We keep the number of canaries small relative to the full training set to avoid interfering with the defense and the primary learning task.

\subsection{Canary construction}\label{sec:canary_const}
The initial canary set is constructed before training the audited model.
Rather than selecting training examples and replacement labels uniformly at random, we use model-informed criteria to choose the examples that are more likely to expose weaknesses.
To obtain these criteria, we use a reference model trained on the original training data before inserting any canaries.
This reference model is used only to select the canary examples and their flipped labels.

\noindent\textbf{Flip-label selection.}\label{sec:lab_select}
For each training example $(\vx,y)$, we use the highest-logit incorrect class under a reference model as the replacement label:
\begin{equation}
    \tilde{y} = \arg\max_{k\in\gY\setminus\{y\}} z_k(\vx),
\end{equation}
where $z_k(\vx)$ denotes the logit for class $k$.
We refer to $\tilde{y}$ as the flipped label.
In the canonical setting, we choose the runner-up class of the reference model as $\tilde{y}$.

\noindent\textbf{Canary selection.}\label{sec:select}
Using $\tilde{y}$, we define the logit margin
\begin{equation}
    m(\vx) = z_y(\vx)-z_{\tilde{y}}(\vx).
\end{equation}
We then approximate the smallest margin attainable within $\epsilon$ using the first-order Taylor expansion:
\begin{equation}
    s_{\mathrm{Taylor}}(\vx) =
    m(\vx) - \epsilon \left\| \nabla_{\vx} m(\vx) \right\|_1.
\end{equation}
A smaller Taylor score indicates that the example is expected to lie closer to a decision boundary under the allowed perturbation budget.
We select $m$ training examples with the lowest Taylor scores as the canaries.

Each selected example $(\vx,y)$ is then replaced by the canary $(\vx,\tilde{y})$ as described in Section~\ref{sec:tt_canary}.

\subsection{Targeted recovery}\label{sec:recovery_audit}
At its core, the audit is a \emph{targeted adversarial example}.
It searches for a perturbation $\delta (\|\delta\|_p \le \epsilon)$ that flips the canary back to the original label $y$:
\begin{equation}
    f(\vx+\delta;\vtheta)=y.
\end{equation}
We obtain the perturbation by solving
\begin{equation}
    \delta^{*}\in
    \arg\min_{\|\delta\|_p \le \epsilon}
    \gL\bigl(z(\vx+\delta;\vtheta),y\bigr).
\end{equation}
The audit is done only for effectively learned canaries (Equation~\ref{eq:learned_canary}).
A successful recovery shows that the behavior learned from the flipped label can be reversed within the perturbation budget of the threat model.
Additionally, we repeat the audit for multiple perturbation budgets to examine how little perturbation is required for recovery.

\subsection{Audit metrics}\label{sec:metric}
We use separate metrics for canary memorization, recovery success, and failure severity.

\noindent\textbf{\Acf{ifr}.}
Before the targeted recovery, we measure how many injected canaries are classified as their flipped label (i.e. effectively learned):
\begin{equation}
    \mathrm{IFR}=
    \frac{1}{|\gC|}
    \sum_{(\vx,y,\tilde{y})\in\gC}
    \mathbbm{1}
    \left[
        f(\vx;\vtheta)=\tilde{y}
    \right].
\end{equation}
\ac{ifr} measures whether the modified label has been learned.
Only the initially flipped canaries are included in the conditional recovery metric below.

\noindent\textbf{\Acf{ctr}.}
We consider only the canaries that are successfully flipped during training:
\begin{equation}
    \gC_{\mathrm{flip}}
    =
    \left\{
        (\vx,y,\tilde{y})\in\gC
        :
        f(\vx;\vtheta)=\tilde{y}
    \right\}.
\end{equation}
For a perturbation budget $\epsilon$, \ac{ctr} is the fraction of initially flipped canaries for which the targeted recovery audit restores the original label:
\begin{equation}
    \mathrm{CTR}(\epsilon)
    =
    \frac{1}{|\gC_{\mathrm{flip}}|}
    \sum_{\substack{
        (\vx,y)\\[-1pt]\in
        \gC_{\mathrm{flip}}
    }}
    \mathbbm{1}\!\left[
        \exists\,\delta:
        \|\delta\|_p\le\epsilon,
        f(\vx+\delta;\vtheta)=y
    \right].
\end{equation}

\noindent\textbf{Empirical recovery radius.}
Severity is measured by the smallest perturbation resulting in a successful recovery:
\begin{equation}
    \rho(\vx)=\inf
    \left\{
        \epsilon :
        \exists\,\delta,\;
        \|\delta\|_p\le\epsilon,\;
        f(\vx+\delta;\vtheta)=y
    \right\}.
\end{equation}
We report this as \ac{mrr}. Smaller recovery radii indicate that the learned flipped behavior can be reversed with a smaller perturbation.

\noindent\textbf{\Acl{asr}.}
For comparison with standard adversarial evaluation, we also report \acf{asr} at a fixed perturbation budget.
Unlike \ac{ctr}, this quantity is conditioned on clean true-positive samples and measures how often correctly classified inputs can be perturbed into targeted incorrect class.

\noindent\textbf{Certified robustness.}
For such defenses, we additionally record the certified radius assigned to each canary and compare it with the radius obtained on regular examples.
A smaller radius indicates a weaker guarantee on the canary, but does not by itself constitute a certificate violation.

\subsection{Adaptive canary refinement}\label{sec:refine}

The initial canaries are constructed without the access to the audited model.
After training, some canaries may lie deep inside the flipped-label region, requiring a large perturbation for recovery. 
Hence, we use the learned decision geometry to move them closer to this boundary while keeping them on the flipped-label side.

Let $\gC^{(0)}$ denote the initial canary set, and let $f_0$ denote the model trained using $\gC^{(0)}$.
For each canary $(\vx_i^{(0)}, y_i, \tilde{y}_i)$, we follow a targeted trajectory toward the original label $y_i$ under $f_0$, subject to
\begin{equation}
    \left\|
        \vx_i^{(1)}-\vx_i^{(0)}
    \right\|_{\infty}
    \le
    \epsilon.
\end{equation}

If the trajectory reaches the decision boundary between $\tilde{y}_i$ and $y_i$, we place $\vx_i^{(1)}$ at the closest point on the $\tilde{y}_i$ side of the boundary.
If the boundary is not reached within $\epsilon$, we move the canary as far as possible toward $y_i$ within the same budget.
If the reference model does not predict $\tilde{y}_i$ at $\vx_i^{(0)}$, we keep the original input unchanged.

The refined set is therefore
\begin{equation}
    \gC^{(1)}
    =
    \left\{
        (\vx_i^{(1)},\tilde{y}_i)
    \right\}_{i=1}^{|\mathcal{C}^{(0)}|}.
\end{equation}
The defended model is then retrained with the refined canary set.
All post-training measurements are computed on the same refined inputs $\vx_i^{(1)}$ that were used during training.

%% file: sec/5_eval.tex
\section{Evaluation}
\label{sec:eval}


\subsection{Experiment setup}
\label{sec:exp_setup}

\noindent\textbf{Main setting.}
Our main experiments use CIFAR-10~\cite{krizhevsky2009learning} with ResNet-18~\cite{he2016deepresiduallearning}.
For empirical defenses, we evaluate Madry \acl{at}~\cite{madry2018deeplearningmodels}, which replaces training examples with adversarial examples generated during training; \Acl{trades}~\cite{zhang2019theoreticallyprincipledtradeoff}, which explicitly balances natural accuracy and robustness through a robustness regularization term; and \Acl{mart}~\cite{wang2020improvingadversarialrobustness}, which emphasizes robustness on misclassified and difficult examples.
For certified defenses, we evaluate \Acf{rs}~\cite{cohen2019certifiedadversarialrobustness}, which provides probabilistic $L_2$ robustness certificates by smoothing predictions with Gaussian noise, and \Acl{crown}-\Acf{ibp}~\cite{zhang2019stableefficienttraining}, which combines interval bound propagation with linear relaxations to certify robustness within a bounded perturbation region.

For each defense, we compare models trained without and with canaries, at canary ratios of $1\%$ and $5\%$.
Unless otherwise specified, we use the canonical canary construction from Section~\ref{sec:canary_const}: Taylor-based selection with the runner-up class as the flipped label.
Each canary replaces its original training example, leaving the input unchanged.

\noindent\textbf{Empirical defenses.}
For Madry \ac{at}, \ac{trades}, and \ac{mart}, we evaluate the audit using a targeted-to-true PGD recovery attack.
For each initially flipped canary $(\vx,y,\tilde y)$, we run 20 projected gradient steps toward $y$, starting from $\delta^{(0)}=0$:
\begin{equation}
\begin{aligned}
    \delta^{(t+1)}
    &=
    \Pi_{\epsilon}\!\left(
        \delta^{(t)}
        -
        \alpha\,\mathrm{sign}\,
        \nabla_{\delta}
    \right.\\[-2pt]
    &\hspace{2.7cm}\left.
        \gL\bigl(z(\vx+\delta^{(t)};\vtheta),y\bigr)
    \right).
\end{aligned}
\end{equation}
The projection constrains the perturbation to the $\ell_\infty$ ball of radius $\epsilon$ around the canary.
Recovery succeeds when $f(\vx+\delta^{(20)};\vtheta)=y$.
We evaluate $\epsilon\in\{1,2,\ldots,8\}/255$ and report fixed-budget results at $\epsilon=8/255$, matching the standard robustness evaluation budget.

\noindent\textbf{Certified defenses.}
We use the same auditing procedure for \ac{rs} and \ac{crown}-\ac{ibp}, while retaining the threat model and certification procedure associated with each defense.
For \ac{rs}, we use smoothing noise $\sigma=0.25$ and evaluate the resulting $\ell_2$ certificates.
For \ac{crown}-\ac{ibp}, we use an $\ell_\infty$ certification radius of $\epsilon=8/255$.
We compare the guarantees obtained on canaries with those obtained on ordinary examples and separately test whether targeted recovery produces a counterexample within a certified region.

\noindent\textbf{Evaluation metrics.}
We use the audit metrics defined in Section~\ref{sec:metric}.
For empirical defenses, we report \ac{ifr} and \ac{ctr} together with the mean empirical recovery radius;
clean and robust accuracy on the test set track changes in conventional model performance.
We additionally report \ac{asr} at $\epsilon=8/255$ as a conventional attack reference.
For certified defenses, we report the certified radii in addition to the canary recovery measurements.


Additional implementation details, defense-specific hyperparameters are provided in the Appendix~\ref{app:implementation}.

\subsection{Auditing empirical robustness}
\label{sec:empirical_audit}

\begin{figure*}[t]
    \centering
    \includegraphics[width=\linewidth]{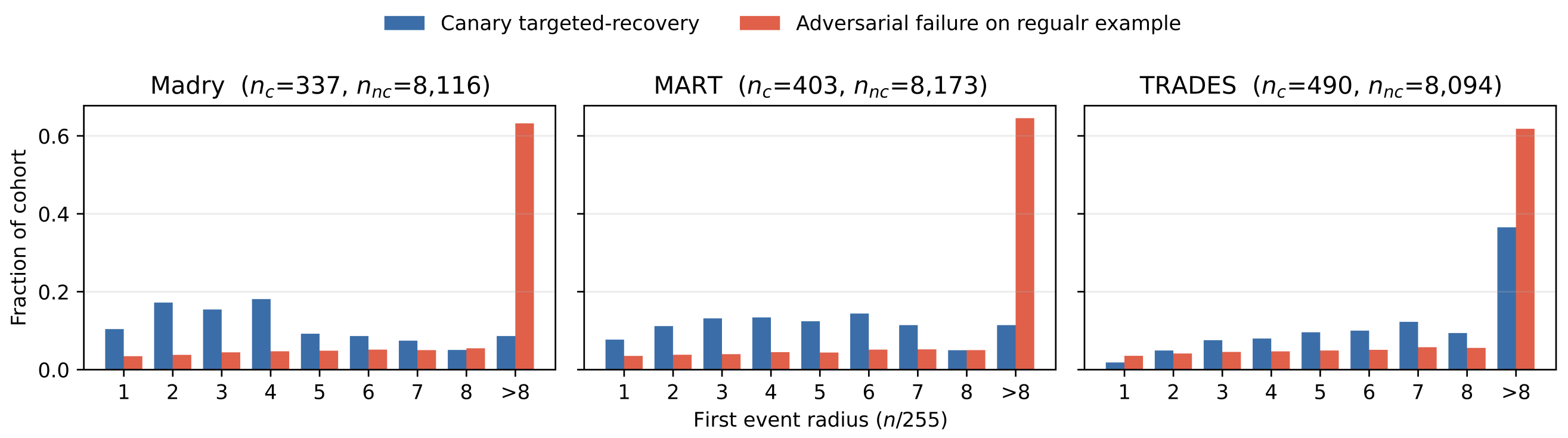}
    \caption{
    Distribution of the first successful attack radius for canary recovery and adversarial failure of regular examples on CIFAR-10/ResNet-18.
    Blue shows targeted recovery of initially flipped canaries; red shows runner-up targeted attacks on regular examples.
    The $>8$ indicates no successful event within $8/255$.
    Canary recoveries generally occur at smaller radii than failures on regular examples.
    }
    \label{fig:recovery-radius}
\end{figure*}

Table~\ref{tab:empirical-main} compares standard robustness evaluation with the canary audit for Madry \ac{at}, \ac{trades}, and \ac{mart}.
Across all three defenses, \ac{ctr} is substantially higher than the conventional \ac{asr} measured under the same evaluation budget.
For example, under Madry \ac{at} with $1\%$ canaries, \ac{asr} is \num{36.82}\%, while \ac{ctr} reaches \num{91.10}\%.
While standard evaluation summarizes average behavior over ordinary samples, the audit concentrates the evaluation on deliberately selected failure-seeking probes.
The large gap between \ac{asr} and \ac{ctr} shows that these probes expose failures that are largely missed by standard average-case evaluation.

\begin{table}[t]
    \centering
    \small
    \setlength{\tabcolsep}{5pt}
    \caption{
    Canary-based auditing of empirical defenses on CIFAR-10 with ResNet-18.
    Rat. denotes the canary ratio.
    Clean and robust accuracy are measured on the standard test set; \ac{asr} measures conventional attack success, \ac{ifr} flipped-label learning, and \ac{ctr} targeted recovery of initially flipped canaries to their original labels.
    Across all three defenses, \ac{ctr} is substantially higher than \ac{asr}, revealing failures that are not apparent from standard aggregate evaluation.
    All values are percentages.
    }
    \label{tab:empirical-main}
    \begin{tabular}{lcccccc}
        \toprule
        Defense & Rat.
        & Clean
        & Robust
        & \ac{asr}
        & \ac{ifr}
        & \ac{ctr} \\
        \midrule
        \multirow{3}{*}{Madry}
            & 0 & 84.22 & 49.43 & 41.31 & -- & -- \\
            & 1 & 81.16 & 51.31 & 36.82 & 67.40 & 91.10 \\
            & 5 & 79.25 & 51.29 & 35.31 & 62.28 & 86.51 \\
        \midrule
        \multirow{3}{*}{MART}
            & 0 & 81.91 & 52.99 & 35.31 & -- & -- \\
            & 1 & 81.73 & 52.72 & 35.49 & 80.60 & 88.59 \\
            & 5 & 78.79 & 52.06 & 33.93 & 74.64 & 77.01 \\
        \midrule
        \multirow{3}{*}{TRADES}
            & 0 & 82.29 & 50.83 & 38.23 & -- & -- \\
            & 1 & 81.16 & 49.79 & 38.65 & 98.00 & 64.29 \\
            & 5 & 79.93 & 49.78 & 37.72 & 94.68 & 61.26 \\
        \bottomrule
    \end{tabular}
\end{table}

The audit also reveals differences between defenses that are less visible in standard metrics.
Robust accuracy and \ac{asr} remain close across Madry, \ac{trades}, and \ac{mart}, whereas \ac{ctr} varies markedly.
This shows how the audit can surface samples exposing a defense’s weakest robustness behavior, even when aggregate metrics appear similar.

Figure~\ref{fig:recovery-radius} compares the first successful attack against canaries vs regular samples.
Across all three defenses, canary recoveries occur substantially earlier and more frequently within the $8/255$ budget, while a large fraction of regular samples remain unbroken throughout the same range.
For Madry \ac{at} with $1\%$ canaries, \num{61.13}\% of memorized canaries are already recovered by $\epsilon=4/255$.
This shows that the audit does not merely find additional failures at the standard evaluation boundary, it exposes weak behavior much earlier within the same perturbation range.

\begin{table*}[t]
    \centering
    \small
    \caption{
    Canary-based auditing of certified defenses on CIFAR-10 with ResNet-18.
    Clean and robust accuracy report the standard evaluation results; \ac{asr} measures conventional attack success, \ac{ifr} flipped-label learning, and \ac{ctr} targeted recovery of initially flipped canaries within the defense-specific evaluation budget -- $\epsilon$ for CROWN-IBP and $\sigma$ for \ac{rs}.
    \ac{ctr}@cert. reports recovery within the corresponding certified radius.
    The radius columns report the mean certified radius for regular and canary examples.
    Across both defenses, canaries receive substantially smaller certified radii than regular examples, while no recovery is observed within the certified regions.
    }
    \label{tab:certified-main}
    \begin{tabular}{llcccccccc}
        \toprule
        Defense & Ratio 
        & Clean
        & Robust
        & \ac{asr}
        & \ac{ifr}
        & \ac{ctr}
        & \ac{ctr}@cert.
        & Regular Rad.
        & Canary Rad. \\
        \midrule
        \multirow{3}{*}{RS}
            & 0\% & 74.41 & 66.07 & 7.45 & -- & -- & -- & 0.455 & -- \\
            & 1\% & 73.38 & 65.19 & 8.25 & 46.80 & 26.92 & 0.00 & 0.449 & 0.0613 \\
            & 5\% & 71.39 & 63.34 & 10.60 & 59.56 & 23.30 & 0.00 & 0.443 & 0.0680 \\
        \midrule
        \multirow{3}{*}{CROWN-IBP}
            & 0\% & 35.94 & 25.82 & 14.80 & -- & -- & -- & 0.0435 &  -- \\
            & 1\% & 34.88 & 26.85 & 15.25 & 32.80 & 31.71 & 0.00 & 0.0459 & 0.0090\\
            & 5\% & 35.26 & 26.35 & 14.29 & 26.16 & 28.29 & 0.00 & 0.0468 & 0.0075\\
        \bottomrule
    \end{tabular}
\end{table*}

\subsection{Auditing certified robustness}
\label{sec:certified_audit}

We next apply the canary audit to two certified defenses, \ac{rs} and \ac{crown}-\ac{ibp}.
Table~\ref{tab:certified-main} reports both the standard robustness measures and the certified radii assigned to canaries and regular examples.

For both defenses, canary insertion causes only modest changes in aggregate robust accuracy, while the certified radii of the canaries are substantially smaller than those of regular examples.
Under \ac{rs} with $1\%$ canaries, the mean certified radius decreases from $0.449$ on regular examples to $0.061$ on canaries.
A similar separation appears for \ac{crown}-\ac{ibp}, where the mean radius drops from $0.0459$ to $0.0090$.
Thus, the audit identifies examples for which the certification procedure provides much weaker local guarantees despite similar aggregate robustness.

We do not observe any targeted recovery within a positive certified radius for either defense.
However, some canaries receive a certified radius of zero, meaning that the certification procedure provides no non-trivial local guarantee for those examples.
This is particularly visible for \ac{crown}-\ac{ibp}, where approximately $4.6\%$ of the audited canaries have certified radius of zero.
Although this does not constitute a certificate violation, it shows that the canary audit can expose regions where the certified guarantee becomes substantially weaker or disappears altogether.

\subsection{Adaptive canary refinement}
\label{sec:refinement}

We evaluate the adaptive refinement procedure introduced in Section~\ref{sec:refine}, which uses the model from the previous training round to reposition canaries before retraining.
Table~\ref{tab:refinement} compares the initial canaries $C^{(0)}$ with two successive refinement rounds for \ac{trades} and \ac{rs}.

Refinement consistently increases \ac{ctr} across both defenses.
For \ac{trades}, \ac{ctr} increases from $64.29\%$ to $73.81\%$ by the second round, while \ac{mrr} decreases from $5.28/255$ to $4.27/255$.
For \ac{rs}, \ac{ctr} increases from $26.92\%$ to $41.75\%$, although the change in \ac{mrr} is small.
This suggests that feedback from the previously trained model can reposition canaries toward regions where targeted recovery is more effective.

However, refinement introduces a trade-off between how reliably the flipped label is learned and how easily that behavior can subsequently be reversed.
The increase in recoverability is comes with a decrease in \ac{ifr}, particularly after the second refinement round.
Clean and robust accuracy remain broadly comparable across rounds, indicating that the observed changes are concentrated on the audit probes rather than coming from worse model performance.

\begin{table}[t]
    \centering
    \small
    \setlength{\tabcolsep}{4.5pt}
    \caption{
    Effect of adaptive canary refinement on CIFAR-10/ ResNet-18 with $1\%$ canonical Taylor canaries.
    $C^{(0)}$ denotes the initial canary set, and $C^{(1)}$ and $C^{(2)}$ successive refinement rounds using the previously trained model.
    \ac{mrr} is the mean first recovery radius under the corresponding threat model.
    Refinement generally increases \ac{ctr} and reduces \ac{mrr}, while clean and robust accuracy remain comparable.
    }
    \label{tab:refinement}
    \begin{tabular}{llccccc}
        \toprule
        Defense
        & Set
        & Clean
        & Robust
        & \ac{ifr}
        & \ac{ctr}
        & \ac{mrr}\\
        \midrule
        \multirow{3}{*}{\ac{trades}}
        &$C^{(0)}$    & 81.16 & 49.79 & 98.00 & 64.29 & 5.28/255\\
        &$C^{(1)}$    & 83.62 & 47.76 & 95.05 & 66.11 & 5.22/255\\
        &$C^{(2)}$    & 81.67 & 47.30 & 87.04 & 73.81 & 4.27/255 \\
        \midrule
        \multirow{3}{*}{\ac{rs}}
        &$C^{(0)}$    & 74.41 & 66.07 & 46.80 & 26.92 & 0.1740\\
        &$C^{(1)}$    & 72.17 & 64.71 & 46.40 & 40.09 & 0.1739\\
        &$C^{(2)}$    & 74.06 & 65.73 & 38.82 & 41.75 & 0.1718 \\
        \bottomrule
    \end{tabular}
\end{table}

\begin{table}[t]
    \centering
    \small
    \setlength{\tabcolsep}{4.5pt}
    \caption{
    Effect of canary selection and flip-label strategy on CIFAR-10/ResNet-18 with \ac{trades} and $1\%$ canaries.
    \ac{asr} and \ac{ctr} are measured under PGD-20 at $\epsilon=8/255$.
    \ac{mrr} denotes the mean recovery radius.
    Taylor-based selection with runner-up labels yields much higher \ac{ifr} while maintaining high \ac{ctr}.
    }
    \label{tab:canary-construction-ablation}
    \begin{tabular}{llcccc}
        \toprule
        Selection & Label
        & \ac{asr}
        & \ac{ifr}
        & \ac{ctr}
        & \ac{mrr} \\
        \midrule

        \multirow{2}{*}{Low-margin}
        & Random    & 38.33 & 18.80 & 55.32 & 4.52/255 \\
        & Runner-up & 38.57 & 23.00 & 62.61 & 4.53/255 \\
        \midrule

        \multirow{2}{*}{Random}
        & Random    & 38.78 & 17.60 & 52.27 & 4.09/255 \\
        & Runner-up & 39.20 & 23.00 & 53.91 & 3.97/255 \\
        \midrule

        \multirow{2}{*}{Taylor-1}
        & Random    & 38.97 & 40.80 & 65.69 & 5.24/255 \\
        & Runner-up & 38.65 & 98.00 & 64.29 & 5.11/255 \\
        \midrule

        \multirow{2}{*}{Taylor-2}
        & Random    & 38.81 & 32.80 & 65.85 & 4.86/255 \\
        & Runner-up & 38.48 & 95.60 & 65.69 & 4.83/255 \\
        \bottomrule
    \end{tabular}
\end{table}

\subsection{Canary construction ablation}
\label{sec:ablation}

We compare alternative canary selection and flip-label strategies in Table~\ref{tab:canary-construction-ablation}. Taylor-based selection yields substantially higher \ac{ifr} than random or low-margin selection, particularly when paired with runner-up labels, while maintaining similarly high \ac{ctr}. This supports the use of Taylor-1 with runner-up labels as our canonical construction.

Low-margin selection prioritizes samples closer to the reference model's decision boundary, while Taylor-2 extends the Taylor-1 criterion with second-order information when ranking candidate canaries. The exact scoring functions and implementation details for low-margin, Taylor-1, and Taylor-2 selection are provided in the Appendix~\ref{app:canary-construction}.

\begin{table*}[t]
    \centering
    \small
    \caption{
    Generalization of of our audit across datasets and architectures using $1\%$ canonical Taylor canaries.
    Madry, MART, and TRADES are evaluated with $L_\infty$ PGD-20 at $\epsilon=8/255$, while RS uses $L_2$ with $\sigma=0.25$.
    In all settings, the audit continues to recover canaries at higher rates than conventional robustness metrics -- the observed audit behavior is not specific to a single dataset or architecture.
    }
    \label{tab:generalization}
    \begin{tabular}{lllcccccc}
        \toprule
        Dataset & Architecture & Defense
        & Clean Acc.
        & Robust Acc.
        & \ac{asr}
        & \ac{ifr}
        & \ac{ctr}
        & Mean Recovery Rad. \\
        \midrule

        \multirow{4}{*}{CIFAR-100}
        & \multirow{4}{*}{ResNet-18}
            & Madry  & 56.26 & 28.24 & 49.66 & 74.00 & 92.16 & 3.72/255 \\
        &   & MART   & 54.03 & 31.80 & 41.14 & 88.80 & 65.09 & 4.61/255 \\
        &   & TRADES & 57.20 & 25.95 & 54.63 & 89.80 & 74.61 & 5.10/255 \\
        &   & RS     & 44.11 & 35.40 & 11.50 & 85.80 & 12.09 & 0.1905 \\
        \midrule
        \multirow{4}{*}{ImageNet}
        & \multirow{4}{*}{ResNet-18}
            & Madry  & 41.60 & 18.54 & 55.43 & 14.65 & 86.95 & 3.33/255 \\
        &   & MART   & 24.77 & 12.76 & 77.73 & 3.01 & 91.71 & 3.43/255 \\
        &   & TRADES & 27.99 & 11.43 & 87.39 & 2.41 & 96.44 & 3.07/255 \\
        &   & RS     & 55.05 & 43.97 & 9.16 & 1.96 & 28.69 & 0.1247 \\
        \midrule
        \multirow{4}{*}{CIFAR-10}
        & \multirow{4}{*}{EfficientNet}
            & Madry  & 56.98 & 37.15 & 35.22 & 31.80 & 93.08 & 3.17/255 \\
        &   & MART   & 65.75 & 43.40 & 34.11 & 52.00 & 95.00 & 2.83/255 \\
        &   & TRADES & 68.46 & 33.55 & 51.02 & 62.40 & 99.04 & 1.87/255 \\
        &   & RS     & 76.60 & 59.64 & 8.65  & 69.60 & 48.85 & 0.2206 \\
        \midrule

        \multirow{4}{*}{CIFAR-10}
        & \multirow{4}{*}{ViT-S}
            & Madry  & 68.81 & 40.16 & 41.67 & 76.60 & 63.71 & 4.40/255 \\
        &   & MART   & 64.70 & 42.52 & 34.33 & 58.40 & 59.93 & 4.74/255 \\
        &   & TRADES & 68.63 & 40.57 & 40.89 & 73.20 & 63.39 & 4.42/255 \\
        &   & RS     & 66.99 & 51.92 & 13.03 & 62.40 & 19.87 & 0.2238 \\
        \bottomrule
    \end{tabular}
\end{table*}

\subsection{Generalization across datasets and models}
\label{sec:generalization}

To test whether the audit behavior extends beyond the default CIFAR-10/ResNet-18 setting, we evaluate additional datasets and architectures, including CIFAR-100~\cite{krizhevsky2009learning} and ImageNet-1k~\cite{deng2009imagenetlargescalehierarchical} with ResNet-18, and EfficientNet~\cite{tan2019efficientnetrethinkingmodel} and ViT-S~\cite{dosovitskiy2020imageworth16x16} on CIFAR-10. As shown in Table~\ref{tab:generalization}, recoverable canary behavior persists across these settings and across both empirical and certified defenses, although its magnitude varies by model and defense.
The audit generalizes across multiple datasets, architectures, and defenses.

\subsection{Audit requirements in practice}
The empirical results satisfy the two primary audit requirements defined in Section~\ref{sec:audit_objective}.
For \ref{req:severity} (\emph{severity}), canary recoveries often occur at substantially smaller perturbation radii than failures on regular examples, showing that the audit can expose particularly weak robustness behavior.
For \ref{req:coverage} (\emph{coverage}), the consistently high \ac{ctr} values across defenses show that these failures are not limited to isolated probes.
Together, these results show that the audit can expose both severe and recurring weak points that are not apparent from standard aggregate evaluation.
\ref{req:cost} (\emph{computational cost}), is discussed separately in the Appendix~\ref{app:compute}.

%% file: sec/6_discussion.tex
\section{Discussion}
\label{sec:discussion}

\noindent\textbf{Algorithm vs data head.}
DP-SGD auditing has two \emph{heads}:
\begin{enumerate*}[label=\arabic*)]
    \item algorithm-level, which constructs the worst case setting to surface the empirical bound of the algorithm or its implementation;
    \item data-level, which is done alongside the regular training procedure.
\end{enumerate*}
\emph{Empirical} robustness methods are limited to the data-level as they are defined w.r.t. a model, dataset and training algorithm.
While certified methods can be audited both ways, e.g., their implementation correctness~\cite{jia2021exploitingverifiedneural}, or empirical behavior (this work).

\noindent\textbf{Attack selection.}
There exists a large body of work showing that adversarial defenses
may not generalize to other $L_p$ norms, attack types or their parametrization; and typically, degrade for $\epsilon$ values higher used in the defense~\cite{carlini2019evaluatingadversarialrobustness,croce2020reliableevaluationadversarial,croce2022adversarialrobustnessmultiple,jiang2024rampboostingadversarial}.
This remains a critical research direction.
On the other hand, we focus on fixed experiment setups (attack, defense, $\epsilon$, norm) and show that existing defense evaluations provide an incomplete depiction of the robustness.

\noindent\textbf{Reference model.}
Our canonical construction uses a reference model to identify informative training examples and replacement labels, but this is not a requirement of the auditing framework.
A fully random construction can be performed without any additional model, as shown in our ablations.
The reference model therefore enables more targeted canary construction at the cost of an additional training run.
Adaptive refinement introduces a similar trade-off by using the model from the previous audit round to reposition canaries, but neither component is required for a basic audit.

%% file: sec/7_related.tex
\section{Related work}
\label{sec:related}

\noindent\textbf{Private training.}
Empirical auditing has been extensively studied for differentially private learning, where attacks give lower bounds on the privacy provided by a training algorithm and detect guarantee violations~\cite{jagielski2020auditingdifferentiallyprivate,tramer2022debuggingdifferentialprivacy,nasr2023tightauditingdifferentially}.
More recent work showed that multiple canary examples can be inserted into the training data to audit DP-SGD with a single training run, substantially reducing the cost~\cite{steinke2023privacyauditingone}.
Recent work studied how the choice of canaries affects auditing power~\cite{boglioni_optimizing_2025}.
Our work draws on this paradigm, but uses training-time canaries to probe adversarial robustness.

\noindent\textbf{Robustness.}
It was shown that numerical errors can lead to violations of certified robustness that an adversary can exploit~\cite{jia2021exploitingverifiedneural}.
Prior work has also shown that adversarial defenses can be circumvented by stronger or defense-aware attacks~\cite{athalye2018obfuscatedgradientsgive,tramer2020adaptiveattacksadversarial,yao2021automateddiscoveryadaptive}.
Other work has shown that adversarial accuracy alone does not fully characterize robustness, as models and individual samples can exhibit different degrees of vulnerability~\cite{olivier_how_2023,han_characterizing_2023}.
In contrast, we keep the defense, attack, and perturbation threat model fixed, and use training-time canaries to expose robustness behavior that conventional evaluation may overlook.

%% file: sec/8_conclusion.tex
\section{Conclusion}
\label{sec:conclusion}

\noindent\textbf{Summary.}
We introduced a canary-based framework for auditing adversarial robustness beyond standard aggregate evaluation.
The framework modifies a small subset of training examples, measures whether the resulting model predicts their flipped labels, and tests whether these predictions can be recovered to the original labels under the same robustness threat model.
Across empirical and certified defenses, we find that canaries can expose weaker robustness behavior than is visible under conventional evaluation, including high targeted recovery rates, failures at smaller perturbation radii, and substantially smaller certified radii.
We also find that defenses with comparable standard robustness metrics can exhibit markedly different behavior under the canary audit.
Our results suggest that active, failure-seeking audits can provide a useful complement to standard robustness evaluation by revealing where robustness is weakest.

\noindent\textbf{Future work.}
We propose a unified auditing framework for empirical and certified robustness,
but one could explore mechanism-specific audits.
Furthermore, our work could be extended to black-box attacks and purification defenses~\cite{nie2022diffusionmodelsadversarial,carlini2022certifiedadversarialrobustness}.
Lastly, our audit aims to showcase worst-case behavior but makes no statement about an average sample, similar to auditing DP-SGD (see Section~\ref{sec:related}).

%% file: sec/9_LLM.tex
\section{LLM Usage Statement}
The authors take full responsibility for all content in this submission.
LLM tools were used to edit and refine author-written text, to generate code, and to help execute experiments; all tool-generated outputs were reviewed and verified by the authors. 
Experimental design, research ideas, claims and conclusions, related work and citations, figures, and data were prepared without LLM use.
The full manuscript was not provided to an LLM provider to obtain reviewer-style feedback.

%% file: sec/X_suppl.tex
\clearpage
\setcounter{page}{1}
\maketitlesupplementary

\appendix
\section{Implementation details}\label{app:implementation}

Code and model checkpoints will be released upon publication.

\subsection{Datasets and architectures} \label{app:d_a}

Our main experiments use CIFAR-10 with ResNet-18.
We additionally evaluate CIFAR-100 and ImageNet-1K with ResNet-18 for dataset generalization, and EfficientNet-B2 and ViT-S/4 on CIFAR-10 for architecture generalization.
CIFAR experiments use standard random cropping with padding and horizontal flipping.
For ImageNet-1K, training images are processed using random resized crops to $224\times224$ and random horizontal flips, while evaluation uses resizing to 256 pixels followed by a $224\times224$ center crop.

\subsection{Canary construction and insertion}\label{app:can_const}

Unless otherwise specified, we use Taylor-1 selection with runner-up label flipping and a canary ratio of $1\%$.
For each defense, a reference model is trained on the unmodified training set and used to construct a defense-specific canary manifest.
Canaries are inserted by replacement: the input remains unchanged, while its original label is replaced by the selected flipped label.
No true-labelled duplicate is retained, so the training-set size remains unchanged.
Canary models are then trained using the same defense-specific recipe as their corresponding baseline models.

\subsection{General setup}

Unless otherwise stated, we follow the publicly available implementations and standard training configurations of the corresponding defenses, including dataset preprocessing, data augmentation, model implementation, optimizer, learning-rate schedule, and training duration.
We retain the original defense objectives and threat-model parameters, and modify only the training data through the canary construction described in Section~\ref{app:can_const}.
We also use the fixed seed for the entire experiments.
Below, we report the settings that are necessary to reproduce our reported experiments.

\subsection{Empirical defense training}

\noindent\textbf{CIFAR-10 with ResNet-18.}
For Madry adversarial training, we train for 150 epochs using SGD with momentum $0.9$, weight decay $5\times10^{-4}$, batch size 128, and initial learning rate $0.1$.
The learning rate is reduced by a factor of $0.1$ every 50 epochs.
The adversarial training inner loop uses $\ell_\infty$ PGD-10 with $\epsilon=8/255$, step size $2/255$, random initialization, and the highest-loss iterate.

MART models are trained for 120 epochs using SGD with momentum $0.9$, weight decay $0.0035$, batch size 128, and initial learning rate $0.01$.
The learning rate is reduced by a factor of $0.1$ at epochs 75, 90, and 100.
The MART inner maximization uses PGD-10 with $\epsilon\simeq8/255$ and step size $0.007$, with $\beta=5$.

TRADES models are trained for 200 epochs using SGD with momentum $0.9$, weight decay $2\times10^{-4}$, batch size 128, and initial learning rate $0.005$.
The learning rate is reduced to $5\times10^{-4}$ at epoch 75, $5\times10^{-5}$ at epoch 90, and $5\times10^{-6}$ after epoch 100.
The TRADES inner maximization uses PGD-10 with $\epsilon=8/255$ and step size $2/255$, with $\beta=6$.

\noindent\textbf{CIFAR-100 with ResNet-18.}
Madry adversarial training runs for 100 epochs using SGD with learning rate $0.1$, momentum $0.9$, weight decay $5\times10^{-4}$, and batch size 128.
The learning rate is reduced by a factor of $0.1$ at epoch 50.
Training uses PGD-10 with $\epsilon=8/255$ and step size $2/255$.

MART is trained for 120 epochs with initial learning rate $0.01$, weight decay $0.0035$, and batch size 128, with learning-rate reductions at epochs 75, 90, and 100.
The inner maximization uses PGD-10 with $\epsilon=8/255$ and step size $2/255$, with $\beta=5$.
TRADES is trained for 100 epochs with batch size 128 and uses PGD-10 with $\epsilon=8/255$, step size $2/255$, and $\beta=6$.

\noindent\textbf{Architecture generalization on CIFAR-10.}
For ViT-S/4, we use AdamW with learning rate $3\times10^{-4}$, weight decay $0.05$, batch size 128, a 10-epoch warm-up, and cosine learning-rate decay.
Adversarial training uses $\ell_\infty$ PGD-10 with $\epsilon=8/255$, step size $2/255$, and random initialization.
MART and TRADES use $\beta=5$ and $\beta=6$, respectively.
The baseline models are trained for 100 epochs; the corresponding canary runs use 70 epochs for MART and 60 epochs for TRADES.
For Madry training, the baseline is trained for 200 epochs and the $1\%$ Taylor-canary model for 100 epochs.

For EfficientNet-B2, Madry $1\%$ and $5\%$ Taylor-canary models are trained for 150 epochs with batch size 256 using SGD with learning rate $0.1$ and weight decay $5\times10^{-4}$.
The learning rate is reduced by a factor of $0.1$ every 50 epochs.
The adversarial inner loop uses PGD-10 with $\epsilon=8/255$, step size $2/255$, and random initialization.
MART and TRADES are initialized from clean ImageNet-pretrained EfficientNet-B2 models and fine-tuned on CIFAR-10 with batch size 128 and learning rate $0.001$.
Both use PGD-10 with $\epsilon=8/255$ and step size $2/255$, with $\beta=5$ for MART and $\beta=6$ for TRADES.
The final canary models are trained using a 30-epoch initial stage followed by 50 additional epochs, for 80 epochs in total.

\noindent\textbf{ImageNet-1K with ResNet-18.}
For MART and TRADES, training uses distributed data parallelism over four GPUs with a global batch size of 512.
We use SGD with momentum $0.9$, weight decay $10^{-4}$, and initial learning rate $0.1$.
The adversarial inner loop uses $\ell_\infty$ PGD-5 with $\epsilon=4/255$ and step size $1/255$.
MART and TRADES use $\beta=5$ and $\beta=6$, respectively, and canary models are trained for 45 epochs.
Robust model selection is performed using PGD-20 validation accuracy.

\subsection{Certified defenses}

\noindent\textbf{CROWN-IBP.}
We evaluate CROWN-IBP on CIFAR-10 using a width-64 ResNet-18 implemented with auto-LiRPA.
Training uses an $\ell_\infty$ target radius of $8/255$ and runs for 200 epochs with batch size 128 using Adam with learning rate $0.001$ and weight decay $2\times10^{-4}$.
The learning rate is reduced by a factor of $0.1$ after $75\%$ of training.
The certification radius is linearly increased from zero to $8/255$ over the first 100 epochs.
The robust-loss weight is $1.0$, and the training objective uses cross-entropy over verified pairwise class margins.

Certification uses CROWN-IBP lower bounds.
An example is certified at a given radius only when all true-vs-other class margin lower bounds are positive.
We evaluate the grid
$\{0,1,\ldots,10,12,14,16\}/255$
and report the largest grid point at which each example remains certified.

\noindent\textbf{Randomized smoothing.}
For CIFAR-10 and CIFAR-100, we use ResNet-18 with Gaussian training noise $\sigma=0.25$.
Models are trained for 90 epochs using SGD with momentum $0.9$, weight decay $10^{-4}$, batch size 256, and initial learning rate $0.1$.
The learning rate is reduced by a factor of $0.1$ at epochs 30 and 60.
Training minimizes standard cross-entropy on inputs perturbed by independent Gaussian noise.

For CIFAR-10 certification, we use $N_0=100$ selection samples and $N=10{,}000$ estimation samples with $\alpha=0.001$.
The certified radius is
$R=\sigma\Phi^{-1}(p_{\mathrm{lower}})$
when $p_{\mathrm{lower}}\geq0.5$; otherwise the classifier abstains.
Certified robust accuracy is measured at an $\ell_2$ radius of $0.25$.
For the CIFAR-100 first-recovery experiments, $N=1{,}000$ estimation samples are used.

For ImageNet-1K randomized smoothing, the ResNet-18 base classifier uses ImageNet normalization with $\sigma=0.25$.
The global certification evaluation uses $N_0=100$ and $N=1{,}000$, while smoothed attack predictions use Monte Carlo voting and EOT for gradient estimation.

\subsection{Attack and evaluation settings}

For CIFAR empirical defenses, standard robust accuracy is measured using untargeted $\ell_\infty$ PGD-20 at $\epsilon=8/255$.
The canary audit uses targeted-to-true PGD-20 and evaluates
$\epsilon\in\{1,2,\ldots,8\}/255$.

The first recovery radius is the smallest evaluated perturbation budget at which an initially flipped canary returns to its original label.
\ac{asr} is conditioned on clean-correct test examples and counts only attacks that change the prediction to an incorrect class.

For the reported ImageNet ResNet-18 empirical models, robust accuracy is evaluated using PGD-20 at $\epsilon=4/255$.
The canary recovery sweep is evaluated separately up to $8/255$.

For randomized smoothing, canary recovery uses targeted-to-true $\ell_2$ PGD-20 with EOT size 8.
CIFAR-10 is evaluated over radii from $0.025$ to $0.25$ in increments of $0.025$, while CIFAR-100 uses $\{0,0.03125,\ldots,0.25\}$.
Abstentions are not counted as attack successes.

For CROWN-IBP, canary recovery uses targeted-to-true PGD-20.
Certificate-conditional recovery attacks each initially flipped canary at its certified flip-label radius.
Test-set \ac{asr} is measured using clean-correct conditioned untargeted PGD-20 at $8/255$.

\subsection{Checkpoint selection}

We select checkpoint with best validation robust accuracy.
Canary-specific metrics such as \ac{ifr}, \ac{ctr}, or recovery radius are not used for checkpoint selection.

\section{Canary construction details}
\label{app:canary-construction}

We construct canaries from training examples that are correctly classified by a reference model.
For each eligible example $(\vx,y)$, we define the runner-up class as the highest-logit class other than the true label,
\[
    r(\vx)
    =
    \arg\max_{k\neq y} z_k(\vx;\vtheta_{\mathrm{ref}}),
\]
and define the corresponding true-versus-runner-up margin as
\[
    m(\vx)
    =
    z_y(\vx;\vtheta_{\mathrm{ref}})
    -
    z_{r(\vx)}(\vx;\vtheta_{\mathrm{ref}}).
\]
The selection rules below rank eligible examples according to different estimates of their proximity to a decision boundary under the audit threat model.

\subsection{Selection criteria}

\noindent\textbf{Random.}
As a model-independent baseline, we uniformly sample from the eligible clean-correct training examples.

\noindent\textbf{Low margin.}
Low-margin selection ranks examples directly by
\begin{equation}
     s_{\mathrm{LM}}(\vx)=m(\vx),   
\end{equation}
and selects examples with the smallest scores.
This prioritizes samples whose true and runner-up logits are already close under the reference model.

\noindent\textbf{First-order Taylor selection.}
Taylor-1 additionally accounts for the local sensitivity of the margin.
For
\begin{equation}
    g(\vx)=\nabla_{\vx}m(\vx),  
\end{equation}
the first-order approximation of the minimum margin within an $\ell_\infty$ ball of radius $\epsilon$ is
\begin{equation}
    s_{\mathrm{T1}}(\vx)
    =
    m(\vx)
    -
    \epsilon
    \left\|
        \nabla_{\vx}m(\vx)
    \right\|_1.    
\end{equation}
We rank candidates in ascending order and select those with the smallest Taylor-1 scores.
For CIFAR experiments, the score is computed in pixel space with $\epsilon=8/255$.

\noindent\textbf{Second-order Taylor selection.}
Taylor-2 extends the same margin approximation with local curvature.
Let
\begin{equation}
    g=\nabla_{\vx}m(\vx),
    \qquad
    H=\nabla_{\vx}^2m(\vx).    
\end{equation}
We approximate the minimum second-order Taylor expansion of the margin as
\begin{equation}
    s_{\mathrm{T2}}(\vx)
    =
    m(\vx)
    +
    g^\top\delta^\star
    +
    \frac{1}{2}
    (\delta^\star)^\top H\delta^\star,
\end{equation}
where
\begin{equation}
    \delta^\star
    \approx
    \arg\min_{\|\delta\|_\infty\leq\epsilon}
    \left[
        g^\top\delta
        +
        \frac{1}{2}\delta^\top H\delta
    \right].
\end{equation}
We do not explicitly construct or store the full Hessian.
Instead, Hessian--vector products $H\delta$ are computed using automatic differentiation.
Starting from $\delta_0=0$, we perform 10 projected sign-descent steps,
\begin{equation}
    \delta_{t+1}
    =
    \Pi_{[-\epsilon,\epsilon]}
    \left(
        \delta_t
        -
        \alpha\,
        \operatorname{sign}
        \left(
            g+H\delta_t
        \right)
    \right),
\end{equation}
with $\epsilon=8/255$ and $\alpha=2/255$.
The final iterate $\delta_{10}$ is used to evaluate $s_{\mathrm{T2}}$, and candidates with the smallest scores are selected.

All CIFAR-10 Taylor scores are computed directly on the \texttt{ToTensor()} representation, so the perturbation radius corresponds to the pixel-space $\ell_\infty$ budget rather than a normalized input space.

\subsection{Flip-label strategies}

Canary selection and flip-label assignment are treated separately.
The Taylor-1 and Taylor-2 scores are always computed using the true-versus-runner-up margin defined above, regardless of the label subsequently assigned to the selected sample.

For \emph{runner-up} flipping, the original label $y$ is replaced by $r(\vx)$.
For \emph{random} flipping, we uniformly sample an incorrect class from the remaining labels.
The image itself is unchanged in both cases.
The original sample is replaced by its relabelled version, so no duplicate with the true label remains in the training set.

\paragraph{Distinction from adaptive refinement.}
Taylor-2 is a pre-training canary \emph{selection} criterion based on a second-order approximation of the reference-model margin.
It is distinct from the adaptive refinement procedure in Section~\ref{sec:refinement}, which starts from an already constructed canary set and uses a canary-trained model to reposition canary inputs before retraining.

\section{Adaptive refinement details}
\label{app:adaptive-refinement}

The main paper describes refinement as an iterative procedure in which information from a previously audited model is used to construct the next canary set.
Here we provide the implementation details.

Starting from a canary set $\gC^{(k)}$, we train a model $f^{(k)}$ and first evaluate all canaries without perturbation.
Only canaries whose prediction equals their flipped training label are eligible for refinement.
For each eligible canary, we use the decision geometry of $f^{(k)}$ to search for an input closer to the decision boundary between the flipped and original labels.

For the empirical $\ell_\infty$ experiments, refinement is constrained to remain within $8/255$ of the current canary input.
The canary is moved toward the relevant decision boundary while remaining on the flipped-label side of that boundary.
When necessary, we use binary search to locate an admissible point near the boundary, terminating the search at a resolution of $1/255$.
If no admissible refinement is found, the canary is left unchanged.

The resulting canary set $\gC^{(k+1)}$ replaces the previous canaries in a fresh copy of the training set, and the defended model is retrained from scratch.
We then re-evaluate the complete inserted canary set, including canaries that were not moved during refinement.
Thus, refinement changes the training probes but does not change the definitions of \ac{ifr}, \ac{ctr}, or recovery radius.

For multiple refinement rounds, all eligibility decisions and refinement directions are recomputed using the model trained in the immediately preceding round.
For example,
\[
    \gC^{(0)}
    \rightarrow
    f^{(0)}
    \rightarrow
    \gC^{(1)}
    \rightarrow
    f^{(1)}
    \rightarrow
    \gC^{(2)}.
\]
The second refinement stage therefore uses $f^{(1)}$ to determine both which canaries remain initially flipped and how they should be moved; it does not reuse the decisions made under $f^{(0)}$.

We additionally consider three refinement variants: refining only canaries for which the boundary search succeeds (\emph{boundary-hit}), refining only the complementary eligible cases (\emph{boundary-miss}), and refining all eligible initially flipped canaries.
These variants isolate how the refinement outcome depends on the initial relationship between a canary and the learned decision boundary.

\section{Additional experiments}
\label{app:additional}

\subsection{Effect of canary ratio}
\label{app:canary-ratio}

\begin{table*}[t]
    \centering
    \small
    \caption{
    Effect of canary ratio on CIFAR-10 with ResNet-18 under Madry AT and MART.
    All canary runs use Taylor-1 selection with runner-up label flipping, and the ratio denotes the fraction of training examples replaced by canaries.
    The results show that injecting more canaries increases the number of available audit probes but does not necessarily strengthen the per-canary audit signal.
    }
    \label{tab:canary-ratio}
    \begin{tabular}{llcccccc}
        \toprule
        Defense & Ratio
        & Clean Acc.
        & Robust Acc.
        & ASR
        & IFR
        & CTR
        & MRR \\
        \midrule
        \multirow{5}{*}{Madry}
            & 0\% & 84.22 & 49.43 & 41.31 & -- & -- & -- \\
            & 1\% & 81.16 & 51.31 & 36.82 & 67.40 & 91.10 & 3.87/255 \\
            & 2\% & 81.62 & 51.13 & 37.32 & 57.73 & 75.59 & 4.25/255 \\
            & 3\% & 81.16 & 51.19 & 38.71 & 67.60 & 79.95 & 4.27/255 \\
            & 5\% & 79.25 & 51.29 & 35.31 & 62.28 & 86.51 & 4.01/255 \\
        \midrule
        \multirow{5}{*}{MART}
            & 0\% & 81.91 & 52.99 & 35.31 & -- & -- & -- \\
            & 1\% & 81.73 & 52.72 & 35.49 & 80.60 & 88.59 & 4.41/255 \\
            & 2\% & 78.21 & 54.06 & 30.94 & 65.80 & 78.88 & 4.63/255 \\
            & 3\% & 78.87 & 53.51 & 32.15 & 76.93 & 69.67 & 4.85/255 \\
            & 5\% & 78.79 & 52.06 & 33.93 & 74.64 & 77.01 & 4.59/255 \\
        \bottomrule
    \end{tabular}
\end{table*}

\begin{table*}[t]
    \centering
    \small
    \caption{
    Ablation of canary construction on CIFAR-10 with ResNet-18 under Madry adversarial training using a $1\%$ canary ratio.
    We vary the canary selection criterion and the incorrect label assigned during training.
    Reference-based selection with runner-up labels yields much higher \ac{ifr} while maintaining high \ac{ctr}.
    }
    \label{tab:canary-construction}
    \begin{tabular}{llcccccc}
        \toprule
        Selection & Flip Label
        & Clean Acc.
        & Robust Acc.
        & ASR
        & IFR
        & CTR
        & MRR \\
        \midrule

        \multirow{2}{*}{Random}
            & Random
            & 82.47 & 51.03 & 38.12 & 3.00 & 80.00 & 3.92/255 \\
            & Runner-up
            & 81.63 & 51.38 & 37.09 & 13.40 & 85.07 & 3.44/255 \\
        \midrule

        \multirow{2}{*}{Low-margin}
            & Random
            & 82.24 & 50.67 & 38.40 & 10.40 & 86.54 & 3.82/255 \\
            & Runner-up
            & 81.78 & 51.62 & 36.88 & 78.00 & 86.67 & 3.94/255 \\
        \midrule

        \multirow{2}{*}{Taylor-1}
            & Random
            & 81.65 & 48.85 & 36.47 & 67.80 & 80.18 & 3.93/255 \\
            & Runner-up
            & 81.16 & 51.31 & 36.82 & 67.40 & 91.10 & 3.87/255 \\
        \midrule

        \multirow{2}{*}{Taylor-2}
            & Random
            & 81.65 & 50.49 & 36.89 & 67.40 & 88.46 & 4.36/255 \\
            & Runner-up
            & 80.68 & 50.95 & 36.37 & 71.20 & 85.17 & 4.07/255 \\

        \bottomrule
    \end{tabular}
\end{table*}

We examine how the scale of the canary intervention affects the audit results while keeping the canary construction fixed in Table~\ref{tab:canary-ratio}.
Across both Madry AT and MART, increasing the canary ratio does not produce a monotonic change in IFR or CTR.
For Madry AT, CTR is highest at $1\%$ ($91.10\%$), decreases at the intermediate ratios, and rises again to $86.51\%$ at $5\%$.
MART shows a similar non-monotonic pattern, with CTR decreasing from $88.59\%$ at $1\%$ to $69.67\%$ at $3\%$, before increasing to $77.01\%$ at $5\%$.
The mean recovery radius remains within a relatively narrow range for both defenses, suggesting that changing the number of injected canaries primarily affects how often the probes are memorized and recovered, rather than substantially shifting the perturbation scale at which successful recovery occurs.

Clean accuracy generally decreases as the intervention becomes larger, whereas robust accuracy remains comparatively stable.
These results indicate that injecting more canaries increases the number of available audit probes but does not necessarily strengthen the per-canary audit signal.


\subsection{Effect of canary construction}
\label{app:exp-canary-construction}

We study how the canary selection criterion and flip-label strategy affect the resulting audit signal, while fixing the canary ratio at $1\%$.
Table~\ref{tab:canary-construction} compares random selection, low-margin selection, and first- and second-order Taylor selection, each with either a random incorrect label or the reference model's runner-up class.

The choice of flip label has a particularly large effect on whether the modified label is learned.
For low-margin canaries, switching from a random label to the runner-up increases IFR from $10.40\%$ to $78.00\%$, while CTR remains nearly unchanged ($86.54\%$ vs.\ $86.67\%$).
Random selection shows the same direction but a smaller change in IFR, from $3.00\%$ to $13.40\%$.
This suggests that assigning a locally plausible alternative class can substantially increase initial flipping, although higher IFR does not by itself imply stronger recovery.

The recovery results depend on the combination of selection and label assignment.
Taylor-1 with the runner-up label achieves the highest CTR at $91.10\%$, compared with $80.18\%$ when the selected examples are assigned random incorrect labels.
Adding second-order information does not further improve the audit: Taylor-2 reaches $88.46\%$ CTR with random labels and $85.17\%$ with runner-up labels.
In particular, the runner-up label is not uniformly beneficial for recovery, despite generally producing higher IFR.

Across all constructions, the mean first-recovery radius remains within a relatively narrow range of $3.44/255$--$4.36/255$.
The construction therefore affects primarily whether a canary adopts the flipped label and whether that flip can subsequently be reversed, rather than substantially changing the perturbation required among successful recoveries.
Overall, Taylor-1 with runner-up flipping provides the strongest audit signal in this comparison, while the second-order selection criterion offers no consistent advantage.

\subsection{Adaptive refinement variants}
\label{app:adaptive-refinement-variants}

In Table~\ref{tab:adaptive-refinement-variants} we evaluate which canaries benefit from adaptive refinement.
Starting from the canary-trained reference model, we identify initially flipped canaries and independently run targeted-to-true PGD-20 from the original canary over a discrete sweep of $\epsilon\in\{1,\ldots,8\}/255$.
We call a canary \emph{boundary-hit} if its prediction remains at the inserted label for at least one nonzero sweep point and then transitions directly to the true label at a larger radius, without first transitioning to a third class.
A \emph{boundary-miss} canary remains at the inserted label throughout the entire sweep.
For refinement, we retain the largest tested perturbation that remains on the inserted-label side: the point immediately preceding the true-label transition for boundary-hit canaries, and the $8/255$ candidate for boundary-miss canaries.
Canaries with a third-class transition, an immediate hit at $1/255$, or no initial flip are left unchanged.

Under this criterion, 490 of the 500 canaries are initially flipped.
We observe a direct transition to the true label for 308 canaries, of which 299 admit a preceding nonzero flip-side point and are therefore refined in the boundary-hit variant.
Another 178 canaries remain at the flipped label throughout the sweep and form the boundary-miss variant.
Four canaries encounter a third-class transition and are excluded from refinement.
We compare refining only the boundary-hit group, only the boundary-miss group, or all eligible canaries.

Refining only one subgroup does not improve recovery over the unrefined audit.
Boundary-hit-only refinement reduces CTR from $64.29\%$ to $58.26\%$, while boundary-miss-only refinement reduces it further to $47.66\%$.
In contrast, refining all eligible canaries increases CTR to $66.11\%$, the highest value among the four settings.
The improvement is therefore not obtained simply by concentrating refinement on canaries for which the reference model already exhibits a direct transition to the true label.

IFR shows a different pattern.
The boundary-miss-only variant has the highest IFR at $98.20\%$ but the lowest CTR, whereas refining all eligible canaries lowers IFR from the unrefined $98.00\%$ to $95.05\%$ while increasing CTR.
Thus, the gain from adaptive refinement cannot be explained by stronger learning of the inserted labels.
Instead, the results indicate that refinement can improve audit recoverability even when the refined canaries are slightly less likely to remain initially flipped after retraining.

The ablation favors refining both boundary-hit and boundary-miss canaries.
Restricting refinement to either subgroup weakens the audit, while applying refinement across all eligible canaries produces a modest improvement over the unrefined baseline.
This also reinforces that IFR and CTR capture different properties of the canaries: a high initial flip rate alone does not imply strong recoverability under the targeted audit.

\begin{table}[t]
    \centering
    \small
    \caption{
    Adaptive refinement variants for TRADES on CIFAR-10 with ResNet-18.
    Boundary-hit and boundary-miss are defined by a targeted-to-true PGD-20 sweep over $\epsilon\in\{1,\ldots,8\}/255$.
    Boundary-hit canaries exhibit a direct transition from the flipped label to the true label after at least one nonzero flip-side point, whereas boundary-miss canaries remain at the flipped label throughout the sweep.
    The benefit of adaptive refinement appears to come from jointly covering heterogeneous canary behaviors, rather than concentrating on a single boundary-defined subset.
    }
    \label{tab:adaptive-refinement-variants}
    \begin{tabular}{lcc}
        \toprule
        Refinement & IFR & CTR \\
        \midrule
        No refinement       & 98.00 & 64.29 \\
        Boundary-hit only   & 92.39 & 58.26 \\
        Boundary-miss only  & 98.20 & 47.66 \\
        All eligible        & 95.05 & \textbf{66.11} \\
        \bottomrule
    \end{tabular}
\end{table}

\begin{table}[t]
    \centering
    \small
    \caption{
    Control evaluation on regular examples for the canonical Madry $1\%$ setting.
    For canaries, Runner-up Rate corresponds to IFR because the reference-model runner-up is the inserted training label.
    For Controls, it denotes the fraction whose clean prediction under the trained model equals their reference-model runner-up; controls are never relabeled during training.
    Regular control examples are also highly recoverable once in the runner-up state, but enter that state much less often than canaries.
    }
    \label{tab:matched-control}
    \begin{tabular}{lcccc}
        \toprule
        Group & $n$ & Runner-up Rate & CTR & MRR \\
        \midrule
        Canary          & 500 & 67.40 & 91.10 & 3.87/255 \\
        Control & 500 & 23.80 & 99.16 & 2.82/255 \\
        \bottomrule
    \end{tabular}
\end{table}

\subsection{Control evaluation on regular examples}
\label{app:matched-control}

\begin{table*}[t]
    \centering
    \small
    \caption{
    ImageNet evaluation for CIFAR-10 TRADES at $\epsilon=8/255$.
    For non-semantic label assignments, Initial State is the fraction predicted as the inserted non-semantic label, and Conditional Success measures targeted recovery to the semantic CIFAR-10 class among that subset.
    For semantic label assignments, Initial State is the fraction predicted as the assigned semantic class, and Conditional Success measures targeted attack success toward a non-semantic reference-model runner-up among that subset.
    External probes behave differently depending on label construction: non-semantic labels support recovery-based auditing, while semantic labels provide a more stable basis for targeted robustness testing.
    }
    \label{tab:external-canaries}
    \begin{tabular}{lllccc}
        \toprule
        Selection & Label & Ratio
        & Initial State
        & Conditional Success
        & Attack Direction \\
        \midrule
        Low-margin & Random-wrong & 1\%
        & 5.40 & 70.37 & Non-semantic $\rightarrow$ Semantic \\
        Low-margin & Runner-up & 1\%
        & 15.20 & 72.37 & Non-semantic $\rightarrow$ Semantic \\
        Near-boundary & Semantic & 1\%
        & 78.80 & 29.19 & Semantic $\rightarrow$ Non-semantic \\
        Safe-inside & Semantic & 1\%
        & 88.20 & 8.16 & Semantic $\rightarrow$ Non-semantic \\
        \midrule
        Low-margin & Random-wrong & 5\%
        & 5.44 & 61.76 & Non-semantic $\rightarrow$ Semantic \\
        Low-margin & Runner-up & 5\%
        & 30.52 & 41.02 & Non-semantic $\rightarrow$ Semantic \\
        Random & Random-wrong & 5\%
        & 6.12 & 60.13 & Non-semantic $\rightarrow$ Semantic \\
        Random & Runner-up & 5\%
        & 33.04 & 38.14 & Non-semantic $\rightarrow$ Semantic \\
        Near-boundary & Semantic & 5\%
        & 85.52 & 10.71 & Semantic $\rightarrow$ Non-semantic \\
        Safe-inside & Semantic & 5\%
        & 88.04 & 13.81 & Semantic $\rightarrow$ Non-semantic \\
        \bottomrule
    \end{tabular}
\end{table*}

Our Taylor-1 canaries are deliberately selected from examples with favorable local decision geometry.
We therefore ask whether the observed recovery behavior is specific to the canary intervention, or whether similarly selected ordinary examples would exhibit the same behavior even without label flipping.
To test this, we construct a matched regular example control set for the canonical Madry $1\%$ setting.

For each canary, we select one ordinary training example from the same true class with the closest Taylor-1 score.
Matching is one-to-one without replacement, and matched controls are never relabeled or otherwise modified during training.
The matched set has the same true-class composition as the canary set, with no overlap between the two groups.
The mean and median absolute Taylor-1 score differences between matched pairs are $0.735$ and $0.587$, respectively.
Both groups are evaluated using the same trained checkpoint and the same targeted-to-true PGD-20 sweep over $\epsilon\in\{1,\ldots,8\}/255$.

In Table~\ref{tab:matched-control} the controls show that highly recoverable examples are not unique to the canary intervention.
Among controls that are already predicted as the reference-model runner-up, $99.16\%$ recover to the true label within $8/255$, compared with $91.10\%$ for the canaries.
The controls also recover at a smaller mean radius, $2.82/255$ versus $3.87/255$.
This suggests that Taylor-1 selection identifies naturally vulnerable examples for which a runner-up-to-true transition is already locally accessible.

The main difference appears in how often the model enters that state.
The trained model predicts the reference-model runner-up for only $23.80\%$ of controls, but for $67.40\%$ of canaries.
Because the two groups are matched by true class and Taylor-1 score, this gap cannot be explained by the selection criterion alone.
The label-flip intervention therefore makes the selected examples substantially more likely to occupy the inserted-label state from which recovery is measured.

This comparison provides a useful interpretation of the canary construction.
Controls indicate that naturally recoverable worst-case examples already exist in the training distribution.
Canary insertion does not need to create recoverability from scratch; instead, it makes the corresponding failure state much more likely to appear after training.
In this sense, the canary acts as a controlled stress-test probe: Taylor-1 selection identifies examples with favorable recovery geometry, while label flipping makes that geometry observable more reliably during the audit.


\subsection{External canary evaluation}
\label{app:external-canaries}

We analyze whether externally sourced ImageNet examples can serve as audit probes for CIFAR-10 TRADES.
External inputs allow two distinct label constructions.
In the non-semantic setting, an image is assigned a CIFAR-10 label that does not match its semantic content, using either a random-wrong or reference-model runner-up label.
In the semantic setting, the assigned CIFAR-10 label matches the image content; for example, an ImageNet \texttt{Pomeranian} image can be mapped to the CIFAR-10 \texttt{dog} class.
These two constructions induce different audit states and are therefore evaluated with different attack directions.

For non-semantic external canaries, we first measure whether the trained model predicts the inserted non-semantic label on the clean image.
Among those initially targeted examples, we run targeted-to-semantic PGD-20 and measure recovery to the semantic CIFAR-10 class.
For semantically aligned external probes, we instead first verify that the assigned semantic label is learned, and then run targeted PGD-20 toward a different non-semantic baseline runner-up class.
All attacks use $\epsilon=8/255$.

\begin{table*}[t]
    \centering
    \small
    \caption{
    Sensitivity to training randomness for the canonical $1\%$ Taylor-1 runner-up canary construction on CIFAR-10 with ResNet-18.
    Results are reported for Madry AT and TRADES across three training seeds.
    The audit metrics remain stable across training seeds, indicating that the observed canary behavior is not driven by a particular random initialization or training trajectory.
    }
    \label{tab:seed-sensitivity}
    \renewcommand{\arraystretch}{1.15}
    \begin{tabular}{llcccccc}
        \toprule
        Defense & Seed
        & Clean Acc.
        & Robust Acc.
        & ASR
        & IFR
        & CTR
        & MRR \\
        \midrule

        \multirow{3}{*}{Madry}
            & 0   & 81.16 & 51.31 & 36.82 & 67.40 & 91.10 & 3.87/255 \\
            & 76  & 80.78 & 51.04 & 36.82 & 63.60 & 88.99 & 3.78/255 \\
            & 137 & 81.29 & 51.01 & 37.26 & 71.00 & 88.45 & 3.87/255 \\
        \midrule

        \multirow{3}{*}{TRADES}
            & 0   & 81.16 & 49.79 & 38.65 & 98.00 & 64.29 & 5.28/255 \\
            & 76  & 81.55 & 50.31 & 38.31 & 97.40 & 60.99 & 5.15/255 \\
            & 137 & 81.66 & 50.25 & 38.46 & 97.20 & 66.46 & 4.95/255 \\
        \bottomrule
    \end{tabular}
\end{table*}

\begin{table*}[t]
    \centering
    \small
    \caption{
    Effect of repeated canary exposure under Madry AT on CIFAR-10 with ResNet-18.
    IN and OUT canaries are constructed using the same Taylor-1 selection rule with runner-up label assignment.
    Only IN canaries are inserted into training, while OUT canaries are excluded.
    Exposure denotes the number of training copies of each IN canary.
    Ordinary controls instead repeat unmodified training examples and therefore have no associated canary metrics.
    Loss-MIA AUROC measures loss-based separability between IN and OUT canaries.
    Increased exposure strengthens memorization but weakens recovery, separating canary memorization from audit recoverability.
    }
    \label{tab:canary-exposure}
    \begin{tabular}{llccccccccc}
        \toprule
        Type & Exposure
        & Clean
        & Robust
        & ASR
        & IN IFR
        & OUT IFR
        & IN CTR
        & OUT CTR
        & Loss-MIA AUROC
        & IN MRR \\
        \midrule

        \multirow{4}{*}{Canary}
        & $0\times$  & 81.05 & 51.19 & 36.85 & 17.0 & 18.6 & 65.88 & 64.52 & 48.70 & 4.48/255 \\
        & $1\times$  & 81.48 & 51.44 & 36.87 & 25.0 & 18.8 & 65.60 & 65.96 & 54.76 & 4.24/255 \\
        & $5\times$  & 80.84 & 49.93 & 38.24 & 45.8 & 19.2 & 47.60 & 56.25 & 70.78 & 5.24/255 \\
        & $10\times$ & 77.49 & 49.46 & 36.21 & 60.2 & 21.2 & 35.88 & 58.49 & 77.74 & 4.54/255 \\
        \midrule

        \multirow{2}{*}{Ordinary}
        & $5\times$  & 82.17 & 50.63 & 38.38 & -- & -- & -- & -- & -- & -- \\
        & $10\times$ & 81.35 & 49.90 & 38.66 & -- & -- & -- & -- & -- & -- \\

        \bottomrule
    \end{tabular}
\end{table*}

Table~\ref{tab:external-canaries} shows that the two label constructions differ most clearly in how reliably they establish the intended pre-attack state.
Non-semantic assignments produce relatively low target occupancy, ranging from $5.40\%$ to $33.04\%$.
Runner-up assignment improves this rate over random-wrong labels, but the inserted state remains difficult to induce on external inputs.
Once that state is reached, however, targeted recovery to the semantic class remains substantial, with conditional recovery between $38.14\%$ and $72.37\%$.
This shows that canary-like recovery behavior can occur outside the original input distribution, although the corresponding target state is less reliably formed.

Semantic assignments show the opposite pattern.
The intended semantic state is learned for $78.80\%$--$88.20\%$ of the external examples, providing a substantially more stable starting point for evaluation.
Among these semantic-learned examples, targeted ASR toward a non-semantic runner-up ranges from $8.16\%$ to $29.19\%$.
Near-boundary selection yields the highest ASR in the $1\%$ setting, but the ordering does not persist at $5\%$, suggesting that boundary proximity is a secondary selection effect rather than a consistent requirement for external vulnerability.

Taken together, the results show that label construction determines the role of an external probe.
Non-semantic assignments provide the closest analogue to the main recovery-based canary audit, but the inserted target state is difficult to induce reliably on OOD inputs.
Semantic assignments instead produce a stable learned state and support a complementary targeted robustness test on semantically valid external examples.
Thus, external inputs can extend the audit beyond the original training distribution, but the appropriate interpretation depends on whether the external label is used to induce a canary-like failure state or to establish a semantic state for subsequent stress testing.


\subsection{Sensitivity to training randomness}
\label{app:seed-sensitivity}

We evaluate whether the observed audit behavior is sensitive to training randomness by repeating the canonical $1\%$ Taylor-1 runner-up setting with three training seeds.
Table~\ref{tab:seed-sensitivity} reports the results for Madry AT and TRADES on CIFAR-10 with ResNet-18.

The audit behavior is stable across seeds for both defenses.
For Madry AT, IFR varies from $63.60\%$ to $71.00\%$, while CTR remains within a narrow range of $88.45$--$91.10\%$.
The corresponding mean recovery radius is also stable, ranging from $3.78/255$ to $3.87/255$.
TRADES shows even less variation in IFR, remaining between $97.20\%$ and $98.00\%$, while CTR ranges from $60.99\%$ to $66.46\%$ and mean recovery from $4.95/255$ to $5.28/255$.
Clean and robust accuracy likewise remain similar across seeds.
These results show that the observed canary behavior is reproducible across training randomness, while allowing for moderate variation in the exact recovery rate.

\begin{table}[t]
    \centering
    \small
    \caption{
    Reference-model sensitivity for the canonical Madry $1\%$ Taylor-1 runner-up construction on CIFAR-10 with ResNet-18.
    The audited-model training seed is fixed, and only the reference model used for canary construction is changed.
    Audit outcomes remain stable across references despite changes in the selected canary set.
    }
    \label{tab:reference-sensitivity}
    \begin{tabular}{lcccccc}
        \toprule
        
        \shortstack{Ref.\\Seed}
        & Clean
        & Robust
        & ASR
        & IFR
        & CTR
        & MRR \\
        \midrule
        0
        & 81.10 & 51.19 & 36.88 & 66.40 & 89.16 & 3.46/255 \\
        76
        & 80.93 & 51.17 & 36.77 & 62.00 & 89.68 & 3.61/255 \\
        137
        & 81.02 & 51.37 & 36.60 & 66.00 & 90.00 & 3.18/255 \\
        \bottomrule
    \end{tabular}
\end{table}

\begin{table}[t]
    \centering
    \small
    \setlength{\tabcolsep}{3.7pt}
    \caption{
    Agreement between reference-model-specific Taylor-1 constructions.
    Shared Canaries reports overlap among the 500 selected examples.
    Runner-up Agreement is measured only on shared canaries.
    Pearson and Spearman correlations are computed over Taylor-1 scores on the common candidate pool.
    Although exact selected sets differ substantially, score geometry and runner-up assignments remain highly consistent.
    }
    \label{tab:reference-agreement}
    \begin{tabular}{lccccc}
        \toprule
        Ref.Pair
        & Shared
        & Jaccard
        & \shortstack{Runner-up\\Agree.}
        & Pearson
        & Spearman \\
        \midrule
        0 vs.\ 76
        & 125/500 & 0.143 & 86.4 & 0.908 & 0.886 \\
        0 vs.\ 137
        & 126/500 & 0.144 & 91.3 & 0.908 & 0.887 \\
        76 vs.\ 137
        & 134/500 & 0.155 & 86.6 & 0.899 & 0.882 \\
        \bottomrule
    \end{tabular}
\end{table}

\begin{table*}[t]
    \centering
    \small
    \caption{
    Computational overhead of the audit on CIFAR-10 with ResNet-18.
    Defense-training time provides the scale of an additional model-training run; reference-based constructions require one such additional run, whereas reference-free constructions do not.
    Manifest construction and evaluation are audit-specific costs.
    All values are completed Slurm wall-clock times.
    }
    \label{tab:computational-cost}
    \begin{tabular}{lllcl}
        \toprule
        Component & Setting & Workload & Wall time & Hardware \\
        \midrule

        \multirow{5}{*}{\shortstack{Defense\\training}}
        & Madry
        & one training run
        & $1{:}22{:}30 \pm 0{:}04{:}21$
        & GH200 $\times$1 \\
        & MART
        & one training run
        & $1{:}27{:}41 \pm 0{:}03{:}19$
        & GH200 $\times$1 \\
        & TRADES
        & one training run
        & $3{:}46{:}48 \pm 0{:}49{:}18$
        & GH200 $\times$1 \\
        & RS
        & one training run
        & $0{:}08{:}04 \pm 0{:}00{:}31$
        & GH200 $\times$1 \\
        & CROWN-IBP
        & one training run
        & $3{:}20{:}50 \pm 0{:}00{:}18$
        & H100 $\times$1 \\

        \midrule

        \multirow{4}{*}{\shortstack{Manifest\\construction}}
        & Random
        & 50k training pool, 500 selected
        & $0{:}19$
        & GH200 $\times$1 \\
        & Low-margin
        & 50k training pool, 500 selected
        & $0{:}19$
        & GH200 $\times$1 \\
        & Taylor-1
        & 50k training pool, 500 selected
        & $0{:}26$
        & GH200 $\times$1 \\
        & Taylor-2
        & 50k training pool, 500 selected
        & $2{:}15$
        & GH200 $\times$1 \\

        \midrule

        \multirow{4}{*}{Evaluation}
        & Empirical
        & 500 canaries, PGD-20, $1$--$8/255$
        & $0{:}36 \pm 0{:}05$
        & GH200 $\times$1 \\
        & Empirical
        & 2,500 canaries, PGD-20, $1$--$8/255$
        & $0{:}57 \pm 0{:}02$
        & GH200 $\times$1 \\
        & RS
        &certification + $\ell_2$ ASR evaluation
        & $47{:}14 \pm 11{:}44$ 
        & GH200 $\times$1 \\
        & CROWN-IBP
        & 2,500 canaries, 14-radius certification
        & $6{:}39$ 
        & H100 $\times$1 \\

        \midrule

        \multirow{7}{*}{\shortstack{Adaptive\\refinement}}
        & TRADES, Round 1
        & refinement construction, 500 canaries
        & $1{:}09$
        & A100 $\times$1 \\
        & TRADES, Round 1
        & retraining
        & $2{:}19{:}01$
        & H100 $\times$1 \\
        & TRADES, Round 2
        & refinement construction + retraining
        & $2{:}14{:}02$
        & H100 $\times$1 \\
        & TRADES, Round 2
        & canary reevaluation
        & $0{:}11$
        & H100 $\times$1 \\
        & RS, Round 1
        & refinement construction, 500 canaries
        & $4{:}54$
        & A100 $\times$1 \\
        & RS, Round 1
        & refinement construction + retraining
        & $9{:}32$
        & H100 $\times$1 \\
        & RS, Round 2
        & refinement construction + retraining
        & $8{:}14$
        & H100 $\times$1 \\

        \bottomrule
    \end{tabular}
\end{table*}
\subsection{Canary exposure and memorization}
\label{app:canary-exposure}

We examine how repeated exposure to the same canary affects memorization and subsequent audit recovery.
We construct matched IN and OUT canaries using the same Taylor-1 selection rule and runner-up label assignment.
Only IN canaries are inserted into training with their flipped labels, while OUT canaries are excluded from training.
We vary the number of IN exposures over $\{0,1,5,10\}$ and evaluate both groups after training.
In addition to IFR and CTR, we report loss-based membership-inference AUROC as an independent measure of memorization.
We also repeat ordinary, unmodified training examples at $5\times$ and $10\times$ exposure to control for the effect of repetition on overall model performance.

Table~\ref{tab:canary-exposure} reveals that repeated exposure substantially increases memorization of the IN canaries.
IN IFR rises from $17.0\%$ at $0\times$ exposure to $60.2\%$ at $10\times$, while the loss-based membership signal increases from an AUROC of $48.70$ to $77.74$.
In contrast, OUT IFR remains between $18.6\%$ and $21.2\%$.
The growing separation between IN and OUT canaries therefore tracks actual training exposure rather than a property shared by examples selected under the same rule.

Stronger memorization, however, does not translate into stronger recovery under the audit.
IN CTR decreases from $65.88\%$ at $0\times$ to $35.88\%$ at $10\times$, despite the simultaneous increase in IFR and membership-inference performance.
OUT CTR remains between $56.25\%$ and $65.96\%$, and the mean first-recovery radius of successfully recovered IN canaries shows no monotonic trend.
Thus, repeated exposure makes the flipped label increasingly likely to be learned, but the canaries that are learned become less likely to recover their original label within the audit budget.

The ordinary-repeat controls help distinguish this behavior from a generic consequence of presenting training examples more often.
At $10\times$, repeating ordinary examples yields $81.35\%$ clean and $49.90\%$ robust accuracy, compared with $77.49\%$ and $49.46\%$ under repeated canary exposure.
The similar robust accuracy but substantially larger clean-accuracy drop for repeated canaries suggests that the aggregate performance change is not explained by repetition alone.
More importantly, the exposure-dependent increase in IFR and loss-based membership signal is specific to the IN canaries.
Overall, these results separate memorization from audit recoverability: stronger exposure produces clearer evidence of memorization, while making the memorized canaries harder, rather than easier, to recover with the targeted audit.

\subsection{Sensitivity to reference model}
\label{app:reference-sensitivity}

Our Taylor-1 canary construction depends on a reference model for both scoring and runner-up label assignment.
We therefore test whether the resulting audit is sensitive to the particular reference-model realization in Table~\ref{tab:reference-sensitivity}.
We fix the audited-model training seed to 0 and vary only the reference-model seed over $\{0,76,137\}$.
For each reference model, we independently recompute Taylor-1 scores, select a new $1\%$ canary set, assign runner-up labels, and train the audited Madry model under the same configuration.

The resulting audit metrics vary little across reference models.
Clean and robust accuracy remain within narrow ranges, while CTR stays between $89.16\%$ and $90.00\%$.
IFR varies more modestly, from $62.00\%$ to $66.40\%$, but the overall recovery behavior remains consistent.
This indicates that the audit outcome is not tied to a particular reference-model seed.

We next examine whether the underlying canary constructions themselves are identical across references in Table~\ref{tab:reference-agreement}.

The exact canary sets overlap only partially, with Jaccard similarity between $0.143$ and $0.155$.
However, Taylor-1 scores are strongly correlated across references, with Pearson correlations around $0.90$ and Spearman correlations around $0.88$.
For canaries selected by both models, runner-up labels also agree in $86.4\%$--$91.3\%$ of cases.
Thus, reference-model variation changes which individual examples are selected more than it changes the underlying local geometry or the resulting audit signal.


\section{Computational costs}
\label{app:compute}

In Table~\ref{tab:computational-cost} we report the computational overhead introduced by the audit relative to training the defense without canaries.
All timings are completed Slurm wall-clock times excluding queueing delays.
Because measurements were collected on different GPU architectures, we report the hardware used but do not compare runtime across defenses.

For a reference-free construction, such as random selection with a random label, adding canaries requires only manifest construction and post-training evaluation beyond the ordinary defense-training run.
For 500 canaries, manifest construction takes 19 seconds and the empirical PGD-20 audit takes $36\pm5$ seconds.
Thus, when no reference model is required, the additional cost of the audit is small relative to model training.

Reference-based constructions additionally require one reference-model training run.
This is the dominant additional cost for low-margin and Taylor-based selection, but it is a one-time cost for a fixed reference checkpoint and can be reused across multiple canary manifests, ratios, label assignments, and audited training runs.
Once the reference model is available, manifest construction remains inexpensive: 19 seconds for low-margin, 26 seconds for Taylor-1, and 2 minutes 15 seconds for Taylor-2.
The additional evaluation cost also scales modestly with the number of canaries, increasing from $36\pm5$ seconds for 500 canaries to $57\pm2$ seconds for 2,500.

Certified defenses retain the same overall accounting: canary insertion does not add another training stage beyond the audited model itself, while the main audit-specific cost is the corresponding certification or attack evaluation.
For randomized smoothing, the combined certification and $\ell_2$ ASR evaluation takes $47{:}14\pm11{:}44$, while the CROWN-IBP certification sweep over 2,500 canaries and 14 radii takes 6 minutes 39 seconds.

Adaptive refinement introduces a larger overhead because each refinement round requires constructing a new canary set and retraining the model.
For TRADES, Round 1 refinement construction takes 1 minute 9 seconds, compared with 2 hours 19 minutes for retraining.
The refinement procedure itself is therefore small relative to the additional training run; the main cost of adaptive refinement comes from repeated retraining rather than from generating the refined canaries.